\documentclass[
11pt,
english,
doublespacing, 
headsepline, 
]{MastersDoctoralThesis} 

\usepackage{threeparttable}
\usepackage{arydshln}
\usepackage{scalerel}
\usepackage{multirow}
\usepackage{hyperref}
\usepackage{amsmath}
\usepackage{amsfonts}
\usepackage[ruled,vlined]{algorithm2e}
\usepackage[capitalise]{cleveref}
\usepackage[utf8x]{inputenc} 
\usepackage[sfdefault]{roboto}  
\usepackage[T1]{fontenc}
\usepackage[printonlyused,withpage]{acronym}
\usepackage{tikz}
\usepackage{pgfplots}
\usepackage{xspace}
\usepackage[autostyle=true]{csquotes} 
\usepackage{aligned-overset}
\usepackage{microtype}
\usepackage{nameref}
\usepackage{pdfpages}
\usepackage{tabularx}

\usepackage{natbib}
\usepackage{bibentry}

\usepackage{threeparttable}

\makeatletter

\newcommand{\labitem}[2]{%
\def\@itemlabel{\textbf{#1}}
\item
\def\@currentlabel{#1}\label{#2}}

\newcommand*{\org@overidelabel}{}
\let\org@overridelabel\@verridelabel
\@ifpackagelater{acronym}{2015/03/21}{
  \renewcommand*{\@verridelabel}[1]{%
    \@bsphack
    \protected@write\@auxout{}{\string\AC@undonewlabel{#1@cref}}%
    \org@overridelabel{#1}%
    \@esphack
  }%
}{
  \renewcommand*{\@verridelabel}[1]{%
    \@bsphack
    \protected@write\@auxout{}{\string\undonewlabel{#1@cref}}%
    \org@overridelabel{#1}%
    \@esphack
  }%
}

\makeatother

\DeclareUnicodeCharacter{2212}{−}
\usepgfplotslibrary{groupplots,dateplot}
\usetikzlibrary{patterns,shapes.arrows,calendar}
\pgfplotsset{compat=newest}

\thesistitle{A Cohesive Distillation Architecture for Neural Language Models} 
\supervisor{Dr. Terry \textsc{Lima-Ruas}} 
\examiner{Prof. Dr.-Ing. Bela \textsc{Gipp}} 
\degree{Master of Science} 
\author{Jan Philip \textsc{Wahle}} 
\addresses{} 

\subject{Biological Sciences} 
\university{\href{https://www.uni-wuppertal.de}{Bergische Universität Wuppertal}} 
\department{\href{http://department.university.com}{Department for Data \& Knowledge Engineering}} 
\group{\href{https://dke.uni-wuppertal.de/}{Data \& Knowledge Engineering Group}} 
\faculty{\href{https://fk6.uni-wuppertal.de}{Faculty of Electrical, Information, and Media Engineering}} 

\AtBeginDocument{
\hypersetup{pdftitle=\ttitle} 
\hypersetup{pdfauthor=\authorname} 
\hypersetup{pdfkeywords=\keywordnames} 
}

\begin{document}

\frontmatter 

\pagestyle{plain} 


\begin{titlepage}
\begin{center}

\begin{center}
    \includegraphics[scale=1.5]{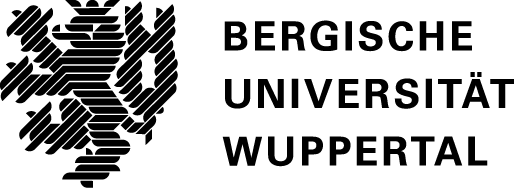}
\end{center}
\vspace{1.5cm}
\textsc{\LARGE Master Thesis}\\[0.5cm] 
\HRule \\[0.4cm] 
{\huge \bfseries \ttitle\par}\vspace{0.4cm} 
\HRule \\[1.5cm] 
 
\begin{minipage}[t]{0.4\textwidth}
\begin{flushleft} \large
\emph{Author:}\\
\authorname 
\end{flushleft}
\end{minipage}
\begin{minipage}[t]{0.4\textwidth}
\begin{flushright} \large
\emph{Examiners:} \\
\examname \\ 
\supname 
\end{flushright}
\end{minipage}\\[2cm]
 
\large \textit{A thesis submitted in fulfillment of the requirements\\ for the degree of \degreename}\\[0.3cm] 
\textit{in the}\\[0.4cm]
\groupname\\\facname\\\univname\\[1.5cm] 

{\large January 27, 2021} 

\vfill
\end{center}
\end{titlepage}

\vspace*{0.2\textheight}

\noindent\enquote{
    \itshape I have no special talents, I am just passionately curious.
}\bigbreak

\hfill Albert Einstein
\vspace*{0.3\textheight}

In the following thesis, the wording ``we'' will be used rather than ``I'' as the ideas were discussed with my advisors and fellow researchers.

\cleardoublepage


\begin{abstract}
    \addchaptertocentry{\abstractname} 
    A recent trend in \acl{NLP} is the exponential growth in \ac{LM} size, which prevents research groups without a necessary hardware infrastructure from taking part in the development process. 
    This study investigates methods for \ac{KD} to provide efficient alternatives to large-scale models. In this context, \ac{KD} means the extraction of information about language encoded in a \acl{NN} and \acl{LKB}.
    
    To test our hypothesis that efficient architectures can gain knowledge from \ac{LM}s and extract valuable information from lexical sources, we developed two methods.
    First, we present a technique to learn confident probability distribution for \acl{MLM} by prediction weighting of multiple teacher networks. Second, we propose a method for \ac{WSD} and lexical \ac{KD} that is general enough to be adapted to many \ac{LM}s.   
    
    Our results show that \ac{KD} with multiple teachers leads to an improved training convergence. When using our lexical pre-training method, \ac{LM} characteristics are not lost, leading to increased performance in \ac{NLU} tasks over the state-of-the-art while adding no parameters. Moreover, the improved semantic understanding of our model increased the task performance beyond \ac{WSD} and \ac{NLU} in a real-problem scenario (\acl{PD}).
    
    This study suggests that sophisticated training methods and network architectures can be superior over scaling trainable parameters. 
    On this basis, we suggest the research area should encourage the development and use of efficient models and rate impacts resulting from growing \ac{LM} size equally against task performance. 
\end{abstract}


\begin{acknowledgements}
\addchaptertocentry{\acknowledgementname} 
\bigskip
\bigskip

\noindent Throughout the writing of this thesis, I have received much support and assistance.

First, I want to thank Professor Bela Gipp for inviting me to complete my graduation at his chair, for his open discussions, and for his support towards my goals.

I am especially grateful to my advisor, Doctor Terry Ruas, for supporting me in all my research aspirations. The last six months have truly been a period of beautiful transformation, and I was lucky to find such a great mentor and friend.

I express my special thanks to Norman Meuschke and Tom{\'a}{\v s} Folt{\'y}nek for their support and helpful discussions in our joint research projects.
Furthermore, I want to thank all Data \& Knowledge Engineering Group members for the valuable discussions and our joint ventures. 

Finally, I want to thank my loving family for their wise counsel and sympathetic ear. Without you, I would not be where I am today. Words simply cannot express how much you mean to me.
\end{acknowledgements}

{
    \hypersetup{linkcolor=black}
    \tableofcontents 
}

{
    \hypersetup{linkcolor=black}
    \listoffigures 
}

{
    \hypersetup{linkcolor=black}
    \listoftables 
}    

\checktoopen
\chapter{List of Abbreviations}
\begin{acronym}[RoBERTa]\itemsep=-2pt
    \acro{ACC}{Accuracy}
    \acro{AE}{Auto Encoding}
    \acro{AR}{Auto Regressive}
    \acro{AI}{Artificial Intelligence}
    \acro{ALBERT}{A Lite BERT}
    \acro{BEM}{Bi-Encoder Model}
    \acro{BERT}{Bidirectional Encoder Representations from Transformers}
    \acro{BoT}{BERT-of-Theseus}
    \acro{BP}{Back Propagation}
    \acro{BPE}{Byte-Pair Encoding}
    \acro{BPTT}{Back Propagation Through Time}
    \acro{CC}{Common Crawl}
    \acro{CNN}{Convolutional Neural Network}
    \acro{CNN}{Convolutional Neural Network}
    \acro{D2V}{Paragraph Vector Model}
    \acro{dEA}{denoising Entity Auto-encoder}
    \acro{ELMo}{Embeddings from Language Models}
    \acro{ERNIE}{Enhanced Language RepresentatioN with Informative Entities}
    \acro{EWISER}{Extended WSD Incorporating Sense Embeddings and Relations}
    \acro{FT}{fastText}
    \acro{GloVe}{Global Vectors}
    \acro{GLU}{Gated Linear Unit}
    \acro{GLUE}{General Language Understanding Evaluation}
    \acro{GPT}{Generative Pre-trained Transformer}
    \acro{GPU}{Graphical Processing Unit}
    \acro{GRU}{Gated Recurrent Unit}
    \acro{KAR}{Knowledge Attention and Recontextualization}
    \acro{KD}{Knowledge Distillation}
    \acro{KL}{Kullback-Leibler}
    \acro{LKB}{Lexical Knowledge Databases}
    \acro{LM}{Language Model}
    \acro{LMMS}{Language Model Makes Sense}
    \acro{LMGC}{Language Model Gloss Classificaiton}
    \acro{LMGC-M}{Language Model Gloss Classificaiton with MLM}
    \acro{LR}{Logistic Regression}
    \acro{LSTM}{Long Short-Term Memory}
    \acro{MC}{Matthews Correlations}
    \acro{MFS}{Most Frequent Sense}
    \acro{ML}{Machine Learning}
    \acro{MLM}{Masked Language Modeling}
    \acro{MPP}{Machine-Paraphrased Plagiarism}
    \acro{MSE}{Mean Squared Error}
    \acro{MSSA}{Most-Suitable Sense Annotation}
    \acro{NB}{Na\"ive Bayes}
    \acro{NLP}{Natural Language Processing}
    \acro{NLU}{Natural Language Understanding}
    \acro{NN}{Neural Network}
    \acro{NSP}{Next Sentence Prediction}
    \acro{PD}{Plagiarism Detection}
    \acro{POS}{Part Of Speech}
    \acro{PPL}{Perplexity}
    \acro{RNN}{Recurrent Neural Networks}
    \acro{RoBERTa}{A Robustly Optimized BERT Pretraining Approach}
    \acro{SC}{Spearman Correlations}
    \acro{SGD}{Stochastic Gradient Descent}
    \acro{SOP}{Sentece Order Prediction}
    \acro{SVD}{Singular Value Decomposition}
    \acro{SVM}{Singular Value Decomposition}
    \acro{W2V}{Word Vector Model}
    \acro{WSD}{Word Sense Disambiguation}
\end{acronym}
~\\
~\\
~\\
~\\
~\\
~\\
~\\






\dedicatory{This thesis is dedicated to my mom for never letting me stop dreaming; to my dad for showing me what really matters in life, and to my sister for always making me laugh.} 


\chapter{Related Publications} \label{AppendixB}
The content included in this master thesis was partially produced in research projects with my fellow researchers. Their respective locations in the master thesis are listed in the following:\\

\nobibliography*
\noindent \acl{WSD} and lexical \acl{KD} methods \cite{Wahle20} (see \Cref{sec:chap3_method_wsd,sec:chap4_eval_wsd}).
\begin{itemize}
    \item[] \bibentry{Wahle20} \\
\end{itemize}
\acl{MPP} detection methods \cite{Wahle21,Wahle21b} (see \Cref{sec:chap3_method_mpp,sec:chap4_eval_mpp}).
\begin{itemize}
    \item[] \bibentry{Wahle21}
    \item[] \bibentry{Wahle21b} \\
\end{itemize}


\mainmatter 

\pagestyle{thesis} 


\acresetall
\chapter{Introduction} \label{Chapter1}

This master thesis addresses a recent \ac{NLP} problem: the increasing \ac{LM} size regarding trainable parameters and computational requirements.
\Cref{Chapter1} introduces the problem of this current trend in language modeling (\Cref{sec:problem_setting}), presents the proposed research objective pursued in this thesis (\Cref{sec:research_objective}), and outlines the remaining chapters (\Cref{sec:thesis_outline}).

\section{Problem Setting and Motivation} \label{sec:problem_setting}
From the first definition of a \ac{LM} for speech recognition systems in the early 1980s \cite{Jelinek97}, LMs have achieved breakthrough results in many \ac{NLU} tasks as diverse as text-summarization, sentiment analysis, or part-of-speech tagging. Since the introduction of the transformer model \cite{VaswaniSPU17}, transfer learning approaches with large-scale pre-trained \ac{LM}s have become a de-facto standard in \ac{NLP} \cite{RadfordWCL18, DevlinCLT19, LiuHCG19}. In September 2020, Google incorporated the large-scale model \acs{BERT} \cite{DevlinCLT19} for almost every English query\footnote{\url{https://tinyurl.com/y6573ufm}}, and Microsoft exclusively licensed\footnote{\url{https://tinyurl.com/y3m7s8zz}} the state-of-the-art model \acs{GPT}-3 \cite{BrownMRS20}.
While these models show significant improvements in downstream tasks, they come at the cost of usually more than a hundred million parameters. Moreover, it appears further increases in the number of parameters, computational budgets, and data often lead to better results \cite{KaplanMHB20}.

\begin{figure}[!hbt] 
    \centering
    \includegraphics[width=\textwidth]{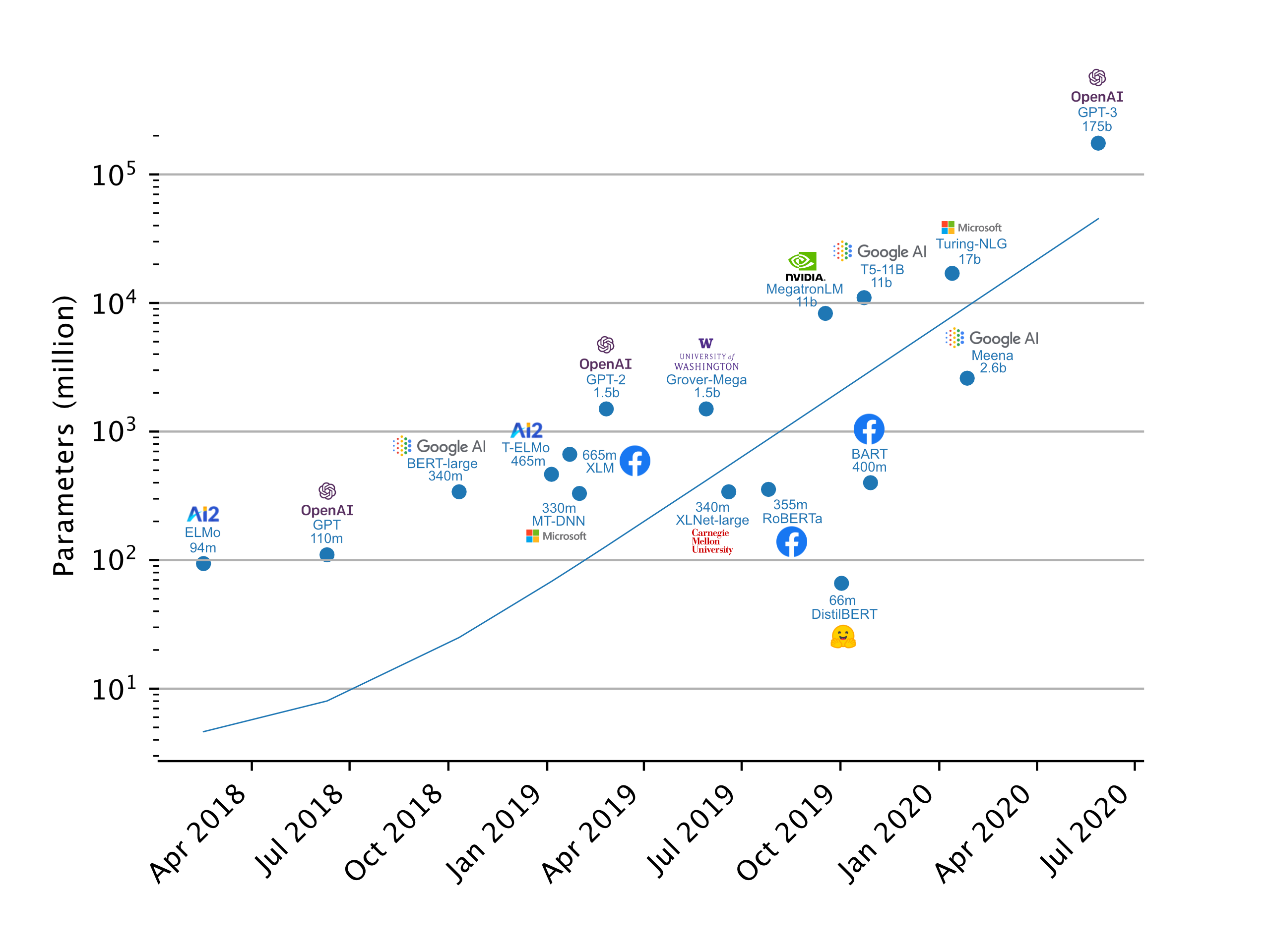}
    \caption[The increase in Language Model parameters over time.]{The increase in Language Model size since the release of ELMo \cite{PetersNIG18} with the number of parameters in logarithmic scale and an exponential approximation of two parameters $e^{0.013x}+3.630$.}
    \label{fig:model_sizes}
\end{figure}

This trend of high increases in model size raises three major concerns.
First, state-of-the-art models are only reproducible by a small group of organizations and institutions with access to the necessary hardware.
\Cref{fig:model_sizes} emphasizes this aspect, where nine large institutions and organizations lead the research with their most impacting models\footnote{Impact refers to the number of citations to the related research paper.}.
\ac{LM} size increasingly prevents researchers without access to the necessary hardware infrastructure from exploring models.
At the time of writing, the largest \ac{LM} contains 175 billion parameters and is only accessible through an official API \cite{BrownMRS20}.
Even if the models were published, researchers would need tremendous amounts of resources to run simple inference in a reasonable time.
Second, energy consumption is in linear relation to model size \cite{SchwartzDSE19}, while model performance seems to follow a logarithmic relation regarding computing budgets \cite{HestnessNAD17, KaplanMHB20}.
The non-linear relation of model performance and energy use makes scaling model size economically and ecologically questionable.
Third, computational and memory requirements prevent the deployment of neural \ac{LM}s on-device and inference in real-time.

Studies show that \ac{LM}s learn redundant features \cite{VoitaTMS19, KovalevaRRR19, MichelLN19}, inspiring the research of alternatives by compression, and \ac{KD} \cite{SanhDCW19, JiaoYSJ20, SunYSL20, GordonDA20}.
Model compression focuses on parameter optimizations and on reducing computational requirements \cite{Tay20}. \ac{KD} leverages the acquired knowledge of large \ac{LM}s into smaller models while reducing their training time \cite{RomeroBKC15, HintonVD15}.
In this work, we use a compressed model that uses its parameters efficiently. We propose two methods for \ac{KD} to extract semantic representations of multiple large \ac{LM}s and to distill the knowledge of \ac{LKB}.
We release the code to reconstruct our experiments and the pre-trained models. Please refer to \Cref{Chapter4} for the corresponding references.
\section{Research Objective} \label{sec:research_objective}
Motivated by the limitations resulting from large \ac{LM}s and the trend of further increasing \ac{LM} size, the following research objective was defined:

\begin{center}
\textbf{\emph{Propose, implement, and evaluate a \ac{LM} training method that distills knowledge from large-scale models and lexical databases into an efficient model to improve its semantic representations.}}
\end{center}

\noindent Therefore, we derived the following research tasks:

\begin{description}
  \labitem{Task 1}{itm:researchtask1} Perform a comprehensive analysis of state-of-the-art \ac{LM}s, their strengths and weaknesses, and methods to distill their knowledge.
  \labitem{Task 2}{itm:researchtask2} Develop and implement a prototype training architecture to incorporate knowledge of larger models and external sources into an efficient architecture.
  \labitem{Task 3}{itm:researchtask3} Evaluate the proposed methods in the tasks of \ac{WSD}, \ac{MPP} detection, and general \ac{NLU} tasks.
\end{description}

\clearpage

\section{Outline} \label{sec:thesis_outline}
\Cref{Chapter1} introduced the problem of current trends in language modeling (\Cref{sec:problem_setting}) and presented the proposed research objective for this thesis (\Cref{sec:research_objective}).

\Cref{Chapter2} addresses research \ref{itm:researchtask1} by providing fundamental background knowledge and analyzing related work in the domain of language modeling (\Cref{sec:chap2_lm}), \ac{WSD} (\Cref{sec:applications_wsd}), and \ac{PD} (\Cref{sec:applications_mpp}). Starting with a definition of \ac{LM}s, we discuss recent models based on the transformer architecture, efficient alternatives, and methods for \ac{KD}.

\Cref{Chapter3} is concerned with research \ref{itm:researchtask2} and presents our developed methods. We begin presenting a novel approach for \ac{KD} with multiple large-scale \ac{LM}s into an efficient architecture (\Cref{sec:multi_teacher}). Next, we propose a training architecture to incorporate knowledge from \ac{LKB} and to perform the task of \ac{WSD} (\Cref{sec:chap3_method_wsd}). We close the chapter by showing how transformer models can be applied to \ac{PD} for machine-paraphrased text (\Cref{sec:chap3_method_mpp}).

\Cref{Chapter4} addresses research \ref{itm:researchtask3} by testing the models' generalization against related publications. We evaluate \ac{WSD} methods incorporating \ac{LKB} on five benchmarks and their generalization on eight \ac{NLU} datasets (\Cref{sec:chap4_eval_wsd}). 
The final experiment evaluates our architecture in the task of \ac{MPP} on a recently presented dataset containing text from research papers, graduation Theses, and encyclopedia articles (\Cref{sec:chap4_eval_mpp}).

\Cref{Chapter5} presents the final considerations to this work. Concluding from the experiments, our methods reduce the amount of computational time and model parameters by using external knowledge and information encoded in large models (\Cref{sec:chap5_conclusion}). We think the presented systems can have a broader impact on other areas (e.g., Mathematical Information Retrieval) and include more researchers in the development process of \ac{LM}s (\Cref{sec:chap5_broader_impact}). Finally, we present research directions for future work (\Cref{sec:chap5_future_work}).

\chapter{Background and Related Work} \label{Chapter2}
\ac{LM}s are a fundamental component of language processing and the main focus of this thesis.
This chapter provides the reader with fundamental background knowledge (\Cref{sec:chap2_fundamentals_lm,sec:chap2_nn_lms,sec:chap2_nn_lms,sec:chap2_rnn_lms}) necessary to explore sophisticated \ac{LM} training architectures based on transformers (\Cref{sec:chap2_transformer,sec:chap2_transformer_lms}).
After introducing efficient architectures for \ac{LM}s (\Cref{sec:chap2_efficient_transformers}), we explore methods to distill the knowledge of a \ac{NN} by comparing three important knowledge concepts (\Cref{sec:chap2_knowledge_concepts}). We review \ac{KD} methods for transformer \ac{LM}s (\Cref{sec:chap2_lm_kd}) and conclude that multi-teacher \ac{KD} and the incorporation of external knowledge appear as promising research directions.
Finally, this chapter presents two applications, the task of \ac{WSD} (\Cref{sec:applications_wsd}) and \ac{MPP} detection (\Cref{sec:applications_mpp}).

\section{Language Models} \label{sec:chap2_lm}
\ac{LM}s assign probabilities to parts of unseen text based on prior knowledge of observed text \cite[Chapter~3]{Jurafsky09}. For example, a \ac{LM} might assign a higher probability to ``a bit of text'' than to ``aw pit tov tags'' because the words in the former phrase occur more frequently in a text corpus \cite{Hiemstra09}.

\subsection{Fundamentals} \label{sec:chap2_fundamentals_lm}
One method to estimate a word's probability in a sentence is by using the \textit{relative count frequency} of sentences \cite[Chapter~3]{Jurafsky09}.
For example, relative count frequency measures the probability of the word ``text'' following the sentence ``a bit of'', by counting occurrences of ``a bit of text'' and relating them to all sentences starting with ``a bit of''. \Cref{eq:lm_count_frequency} formalizes the conditional probability expressed by relative count frequencies (where $P$ is a probability measure and $C$ is a counting function over the corpus).

\begin{align} \label{eq:lm_count_frequency}
    P(\textnormal{``text''} \mid \textnormal{``a bit of''}) &= \frac{C(\textnormal{``a bit of text''})}{C(\textnormal{``a bit of''})}
\end{align}

While estimating probabilities from word counts works well in many cases, the corpus size has to be large and unseen variations (e.g., "A bit of \textit{fun}") have no assigned probability.
Also, modeling the joint probability of a sentence requires estimating all sentences' probabilities with the same number of words. The poor generalization resulting from sparse representations with many zero probabilities and a high modeling complexity inspired more sophisticated methods for modeling the joint probability of words. Before continuing with these methods, we define a standard set of terms and symbols.

We represent a string of contiguous characters as a \textit{token} and an algorithm that transforms a sequence of characters into tokens as a \textit{tokenizer}. In this work's context, a \textit{token} represents either a word or a word piece (i.e., a sub-word which, together with other sub-words, can construct a word). We use the terms \textit{token} and \textit{word} interchangeably. A popular technique for tokenization in the area of \ac{LM}s \cite{RadfordWCL18, DevlinCLT19} is the compression algorithm \ac{BPE} \cite{SennrichHB16} that many tokenizers use, e.g., SentencePiece \cite{KudoR18} or WordPiece \cite{SchusterN12}. \ac{BPE} initializes a set of every character in a corpus to learn efficient merging rules. The algorithm finds frequent (sub-)words and encodes them efficiently using Huffman coding \cite{Furht06}.

We represent a sequence of $n$ words as $w_1, ..., w_n$ or $w_{\leq n}$ (with $w_{<n}$ meaning $w_1,...,w_{n-1}$). The joint probability of words in a sequence having a particular value is $P(w_1, ..., w_n)$. With this notation, we can model the joint probability of a sequence as a forward product by using the chain rule (see \Cref{eq:lm_joint_probability}).

\begin{align} \label{eq:lm_joint_probability}
    P(w_1,...,w_n) &= P(w_1) P(w_2 \mid w_1) P(w_3 \mid w_{<3})P(w_n, w_{<n}) \\
    &=\prod_{i=1}^n P(w_i \mid w_{<i})
\end{align}

\ac{LM}s can be generally divided into \ac{AR} and \ac{AE} models. \ac{AR} models use the forward product of \Cref{eq:lm_joint_probability}, a backward product or a combination of both \cite{PetersNIG18}. A backward product is analogous to the forward one in but with right-to-left context, and the factorization $P(w1,...,w_n) = \prod_{i=n}^1 P(w_i|w_{>i})$.
\ac{AE} models do not perform explicit density estimation but reconstruct corrupted inputs.

\ac{AR} language models can gain robustness by applying the memoryless Markov property \cite[Chapter~4]{Markov62} to the conditional probability estimation. A \textit{bigram} models words' probabilities depending to one previous word, i.e., $P(w_i \mid w_{<i}) \approx P(w_i \mid w_{i-1})$. Bigrams reduce zero probabilities for unseen text as word pairs can model a large fraction of possible sentences. For example, when the bigrams ``a bit'', ``bit of'', and ``of fun'' occured in the corupus, the probability $P(\textnormal{``a bit of fun''})$ can be modeled although this exact phrase never occured.
A generalization of bigrams are \textit{n-grams} which access $n-1$ previous words from the context. Smoothing techniques for n-grams \cite{KneserN95, ChenG96} eliminate the problem of zero probabilities by reallocating the probability mass of frequent and infrequent n-grams.

\subsection{Neural Network Language Models} \label{sec:chap2_nn_lms}

A profound downside of n-gram \ac{LM}s is \textit{the curse of dimensionality} that results from modeling the joint probability distribution for all possible n-grams over the \textit{vocabulary} $V$, which is the set of all (sub-)words in a corpus. The number of possible n-grams grows with $|V|^n$, making the number of parameters to represent large vocabularies and long n-grams infeasibly large. To overcome the curse of dimensionality and to capture long contexts without zero probabilities, \newcite{Xu00} first introduced a feed-forward \ac{NN} to learn a bigram model and \newcite{BengioDVJ03} used a \ac{NN} to directly approximate each conditional probability distribution from \Cref{eq:lm_joint_probability}. The \ac{NN} receives context words $w_{<i}$ as inputs, which are transformed from their one-hot encoding into a continuous feature vector space by a projection matrix $C \in \mathbb{R}^{|V| \times h}$, with $h$ being the \ac{NN}s hidden dimensions to learn each conditional distribution. The model predicts $P(w_i \mid w_{<i})$ using the softmax function \cite[Chapter~6]{Goodfellow16} and optimizes its weights using negative log-likelihood (corresponding to the multi-class cross-entropy \cite[Chapter~4]{Bishop06}), the \ac{BP} algorithm \cite{RumelhartHW86}, and \ac{SGD} \cite{KieferW52}.

Bi-products of performing language modeling with \ac{NN}s are high-level semantic features produced in hidden layers with increasing abstractions towards the prediction \cite{BengioCV13}.
Hidden representations are often referred to as word embeddings and used in many \ac{NLP} tasks \cite{WangWCW19}. Compared to \textit{sparse} vectors capturing word frequencies with integers often including many zeros, \ac{NN} embeddings are \textit{dense} as each value is a real number which results in higher modeling capabilities when the vector size remains equal.
Static word embedding methods \cite{MikolovCCD13, PenningtonSM14, BojanowskiGJM17} use a unique fixed vector to represent a word's meaning. Context-aware word embedding models \cite{PetersNIG18} encode a word with different dense vectors if the context changes.

\subsection{Recurrent Neural Network Language Models} \label{sec:chap2_rnn_lms}
Using feed-forward \ac{NN}s as the conditional probability approximator for language modeling has two significant drawbacks.
First, fully-connected layers expect a fixed-sized input, making the context size fixed too. Second, feed-forward layers treat inputs as independent and simultaneous features.
However, in practice, the context size of sentences varies, and word order can change sentences' semantic content. 
To introduce higher-level features by capturing longer contextual information and processing variable sized contexts, \newcite{Mikolov2011a, Mikolov2011b} used \ac{RNN}s \cite{RumelhartHW86} as function approximators for neural \ac{LM}s.
As traditional \ac{RNN}s tend to be less stable due to the vanishing and exploding gradient problem \cite{Hochreiter91, Bengio93, Bengio94}, \ac{LSTM} networks \cite{Hochreiter97} gained attention for neural \ac{LM}s \cite{Sundermeyer12}. 

An extension to \ac{LSTM}s are bidirectional \ac{LSTM}s \cite{PetersNIG18, HowardR18} which use an additional backward product, analogous to the forward one in \Cref{eq:lm_joint_probability} with the factorization $P(w1,...,w_n) = \prod_{i=n}^1 P(w_i|w_{>i})$.
Other network architectures include \ac{GRU} and \ac{CNN}.
When processing character-level information, word embedding models \cite{PetersNIG18, Ma20} sometimes use \ac{GRU} networks \cite{ChovGB14} to construct contextual character embeddings serving as inputs to the model.
\ac{CNN}s, the most used architecture in computer vision tasks \cite{LeCunBH15}, mainly perform classification tasks in \ac{NLP}, e.g., sentence matching \cite{Hu14}, topic categorization \cite{Kalchbrenner14}, or relation extraction \cite{Nguyen15}, but are not popular in the language modeling domain \cite{Pham16}.

\subsection{Attention in Recurrent Neural Networks} \label{sec:chap2_attention_rnn}
When modeling the probability of a word using its context, intuitively, not every surrounding word is relevant. 
For example, modeling the probability of the word ``python'' given a context requires a specific focus on related words such as ``snake'' or ``programming language''.
This mechanism is a well known cognitive concept studied in psychology called \textit{attention} \cite[Chapter~3]{Anderson15}. Our brain selectively concentrates on a specific stimulus to solve a problem while ignoring other perceivable stimuli.
Attention for \ac{RNN}s and \ac{LSTM}s originated in sequence-to-sequence models, where the model generates one word at a time from an input sequence of words \cite{Bahdanau16}.
Focussing on specific context words while ignoring less important ones, the attention mechanism uses learnable coefficients \cite{Tran16,Mei16}.

Although \ac{RNN}s and \ac{LSTM}s using attention show superior performance in many \ac{NLP} tasks, their recurrent dependence requires a sequential computation of gradients within its layers from the last word to the first one \cite[Chapter~10]{Goodfellow16} (sometimes referred to as \ac{BPTT}). \ac{BPTT} prevents \ac{RNN}s and their derivatives to scale with more hardware as sequential parts cannot be parallelized. Modern neural \ac{LM}s leverage a variation of attention and feed-forward \ac{NN}s that are highly parallelizable and discussed in the following Section.

\subsection{The Transformer Model} \label{sec:chap2_transformer}
The introduction of the transformer model \cite{VaswaniSPU17} for neural machine translation revolutionized neural language modeling \cite{RadfordWCL18, DevlinCLT19, LiuHCG19}. 
Transformers are feed-forward \ac{NN}s using an encoder-decoder structure. The encoder maps a sequence of token embeddings to a continuous representation, which is forwarded to the decoder generating an output sequence one word at a time.
As the transformer uses feed-forward layers, it overcomes vanishing gradient problems that make learning long-term dependencies difficult and allows for high parallelization, reducing training time. \newcite{VaswaniSPU17} further improve long-term dependency learning by a novel self-attention mechanism that considers the connection of every word to every other word in the input sequence.

Encoder and decoder consist of consecutive modules. The output representations of the final encoder state serve as the input to each decoder module.
Each processing step involves mainly three components: fully-connected layers, residual connections, and multi-headed self-attention.
Fully-connected layers project the embedding space into a higher dimension to construct more complex features. Residual connections \cite{HeZRS15} increase the gradient signal for \ac{BP} by forwarding the identity of the previous layer to the next one achieving higher generalization than chain-like networks \cite{He20}.
The most effective and a well studied architectural component of transformers \cite{KitaevKL20,BeltagyPC20} is the scaled dot-product attention (which we will refer to as self-attention).

Self-attention estimates a probability distribution over word co-relations within a sequence using softmax. Therefore, the transformer learns three weight matrices, the Key ($K$), Query ($Q$), and Value ($V$). Each matrix is a linear transformation of the word embeddings and has size $s \times d_k$, where $s$ is the sequence length, and $d_k$ is the hidden dimension of the $K$ matrix. Self-attention uses the dot product of $Q$ and $K$ as a similarity measure between word representations. The softmax function applied to the dot product of $Q$ and $K$ yields an attention score, i.e., a probability distribution representing all word-pairs' influence over the sequence. An additional normalization term prior to the softmax ($\sqrt{d_k}$) reduces large values causing diminishing gradients.
The probability distribution multiplied by $V$ selects word representations in proportion to word co-relateions (see \Cref{eq:lm_attention}). 

\begin{align} \label{eq:lm_attention}
    Att(Q, K, V) &= \textnormal{softmax}\left(\frac{QK^T}{\sqrt{h}}\right)V
\end{align}

Compared to \ac{RNN}s, self-attention and fully-connected layers are large matrices products and can be highly parallelized. Therefore, transformers are scalable and train in a fraction of the time compared to \ac{RNN}s with the same number of floating-point operations. However, as transformers are feed-forward, they have no time-step dependence like \ac{RNN}s and require positional information in the input to encode its token position. Transformers typically use a combination of sine and cosine functions of different frequencies added to the token embeddings to represent the word position in the sequence.

\begin{figure}[htb]
    \centering
    \includegraphics[width=.8\textwidth]{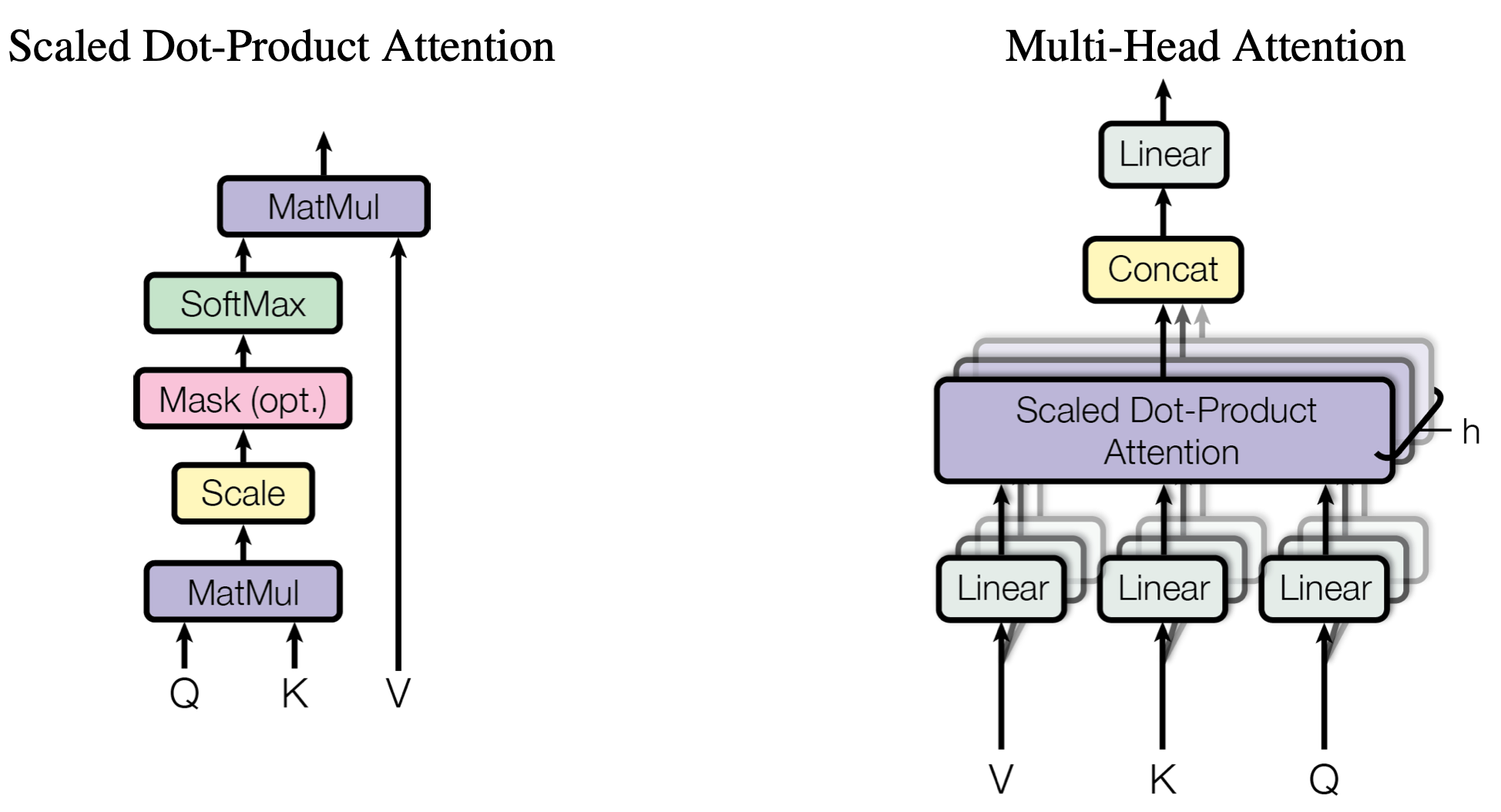}
    \caption[Scaled Dot-Product Attention and Multi-Head Attention.]{Scaled Dot-Product Attention (left). Multi-Head Attention (right) \cite{VaswaniSPU17}.}
    \label{fig:attention}
\end{figure}

In an attempt to capture different aspects of a language, multi-headed self-attention stacks multiple self-attention layers in parallel, similar to kernels in a \ac{CNN}. Multi-headed self-attention allows the transformer to learn multiple features in each layer. Both methods are illustrated in \Cref{fig:attention}.

Self-attention is an effective method to perform neural machine translation, creating semantic encoder representations. These high-level representations inspired many works in the domain of language modeling \cite{RadfordWCL18,DevlinCLT19,LanCGG19,BrownMRS20} and shifted the paradigm of how language modeling is approached (see \Cref{sec:chap2_transformer_lms}).

\subsection{Transformer-Based Language Models} \label{sec:chap2_transformer_lms}
Universal \ac{LM} fine-tuning \cite{HowardR18} introduced an inductive transfer learning paradigm for \ac{LM}s.
Their model first optimizes for the conditional probability distributions over large corpora to capture general language features (pre-training) and then learns new tasks with few additional gradient steps on new datasets (fine-tuning). Pre-training is expensive and typically requires multiple billion tokens but is performed only once. Fine-tuning is cheaper and processes a fraction of the tokens used in pre-training.
\newcite{HowardR18} inspired many models to follow the transfer learning paradigm \cite{RadfordWCL18, DevlinCLT19}.

Using the transformer architecture, \ac{BERT} \cite{DevlinCLT19} proposes two pre-training tasks to capture general language aspects, i.e. \ac{MLM} and \ac{NSP}.
\ac{MLM} is an \ac{AE} training method as it reconstructs masked words within a context. For example, given the sentence ``I use the programming language python'', words are masked at a certain probability using a special token. The sentence may transform into ``I \texttt{[MASK]} the programming \texttt{[MASK]} python'' challenging \ac{BERT} to reconstruct the masked words using a bidirectional context.
\ac{NSP} predicts whether two sentences are semantically connected. Multiple studies showed the \ac{NSP} task has little influence on \ac{BERT}'s performance \cite{LanCGG19, LiuHCG19} which is why we focus on the \ac{MLM} task in our models (\Cref{sec:chap3_method_wsd,sec:multi_teacher}). For more information about the \ac{NSP} task, we suggest the original work of \newcite{DevlinCLT19}.
\ac{BERT} transforms the final representation of the transformer for each masked word into a probability vector over the vocabulary using softmax and optimizes for the targets using cross-entropy loss.

The same way \ac{W2V} \cite{MikolovSCC13} inspired many models in \ac{NLP} \cite{BojanowskiGJM17,RuasGA19,RuasFGd20}, \ac{BERT} echoed in the literature with recent models as well \cite{YangDYC19, ClarkLLM20}.
\ac{RoBERTa} \cite{LiuOGD19} showed additional training data from, a more extensive vocabulary, and more training steps with larger batches, further improve \ac{BERT}'s performance. 

XLNet \cite{YangDYC19}, an extension of pre-training methods from Transformer-XL \cite{DaiYYC19}, aims to take advantage of \ac{AE} concepts for \ac{AR} language modeling. XLNet explores two of \ac{BERT}'s deficiencies: (1) the corruption of sequences with artificial \texttt{[MASK]} tokens, which never occur in regular text, and (2) the prediction of masked words in a single step, assuming independence of words from each other. XLNet removes \texttt{[MASK]} tokens and maximizes the expected log-likelihood of a sequence concerning all possible permutations of the factorization order.

ELECTRA \cite{ClarkLLM20} changes the \ac{MLM} task of \ac{BERT} to a generator-discriminator setup. The model substitutes tokens with artificially generated ones from a small masked \ac{LM} and discriminates them in a noise contrastive learning process \cite{GutmannH10}. BART \cite{LewisLGG19} pre-trains a bidirectional \ac{AE} and \ac{AR} transformer in a joint structure. A two-stage denoising \ac{AE} first corrupts the input with an arbitrary function (bidirectional) and uses sequence-to-sequence to reconstruct the original input (\ac{AR}).

Other popular examples of \ac{AR} models include the three versions of the \ac{GPT} model, with \ac{GPT}-3 reaching 175 billion parameters. The training method of \ac{GPT}-3 mainly relies on predicting consecutive tokens and a large carefully pre-processed dataset composed of \ac{CC} web documents\footnote{\url{http://commoncrawl.org/}}.
The T5 model formulates transfer learning tasks as textual questions and performs prediction using text generation of a large \ac{AR} model.
In this work, we exclude the exploration of pure \ac{AR} models for our models as the targeted downstream tasks do not require text generation. Instead, we focus on the variation of \ac{BERT}-related \ac{AE} models which are compatible with each other for \ac{KD}.

Since the introduction of \ac{BERT}, a growing area of research is concerned with the question of how to reduce computational requirements and parameter count of \ac{LM}s without significant performance loss \cite{Tay20}.
DistilBERT \cite{SanhDCW19} uses the concept of \ac{KD} \cite{HintonVD15} with a student-teacher architecture to extract \ac{BERT}'s knowledge into a smaller model. Therefore, DistilBERT initializes with a selection of \ac{BERT}'s layers and optimizes the negative log-likelihood for soft target probabilities, i.e., \ac{BERT}'s logits, with regular introductions of ground truth labels. \ac{KD} requires fewer training steps compared to training the model from scratch with only ground truth targets and converges to smaller validation losses.

\ac{ALBERT} \cite{LanCGG19} makes \ac{BERT} more efficient through factorized embedding parameterizations, cross-layer parameter sharing, and \ac{SOP}. Compared to \ac{BERT}, RoBERTa, and XLNet which set the hidden layer size $h$ to the WordPiece Embedding size $E$, i.e. $H=E$, \ac{ALBERT} minimizes computational requirements for typically large vocabularies $V$ from $\mathcal{O}(V \times H)$ to $\mathcal{O}(V \times E + E \times H)$ with small values for $E$. \ac{ALBERT} shares all parameters across layers to reduce parameter count significantly (10\% of \ac{BERT}'s parameters for the base model), similar to other strategies for transformers \cite{Dehghani18, Bai19}. \ac{ALBERT} proposes \ac{SOP}, as RoBERTa and XLNet found \ac{NSP}'s impact unreliable due to low task difficulty. \ac{SOP} uses as positive examples two consecutive sentences from the same document, and as negative examples the same sentences but with their order swapped making the task more difficult.

The presented transformer-based \ac{LM}s rely on self-attention, which is the most computationally expensive layer in the transformer's \ac{NN}, growing quadratically concerning the sequence length.
As one example of exploring new schemes to calculate self-attention with the motivation for capturing larger contexts, Longformer \cite{BeltagyPC20} combines a windowed local-global self-attention scaling linearly with the sequence length in comparison to other models. \Cref{sec:chap2_efficient_transformers} introduces a taxonomy for efficient transformer architectures which are relevant to explore large-scale \ac{LM} distillation.

\subsection{Efficient Transformers} \label{sec:chap2_efficient_transformers}
In the first formalization of intelligent machines, Alan Turing described a phenomenon of \ac{ML} with remarkable precision: \\

\enquote{\itshape [M]achines of this character can behave in a very complicated manner when the number of units is large.}\bigbreak

\hfill Alan Turing (1948) ``Intelligent Machines'', page 6 \\

\begin{figure}[htb]
    \centering
    \includegraphics[width=.9\textwidth]{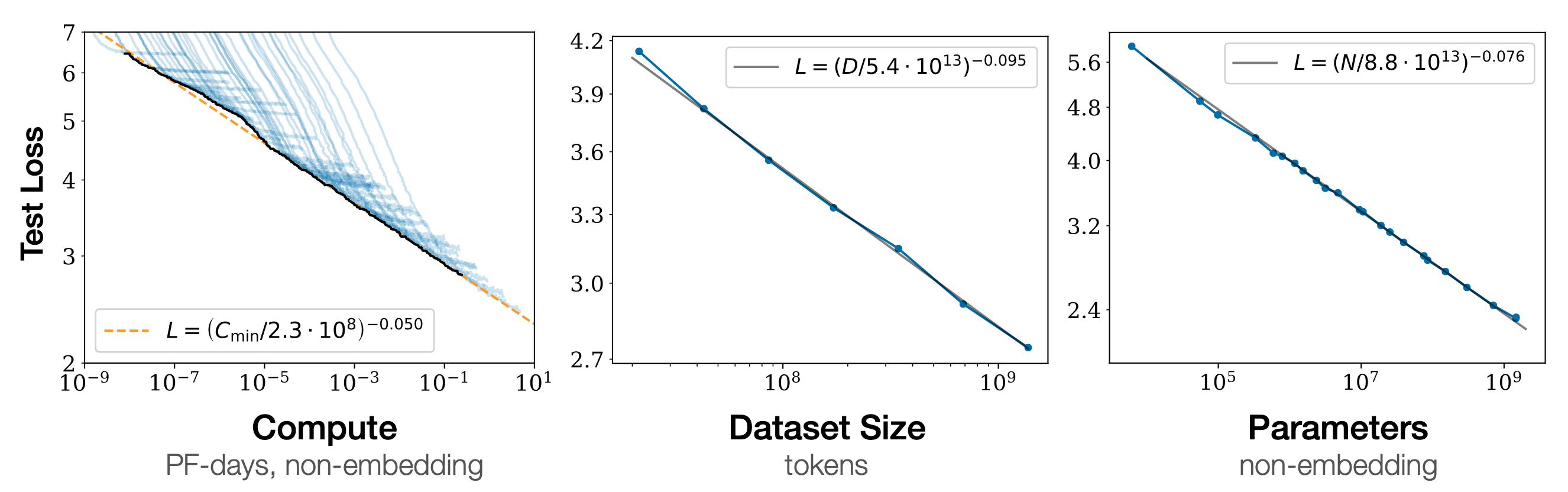}
    \caption[Test loss for transformer models with varying hyperparameters.]{Test loss for transformer language models with varying compute budgets in Peta Flops (PF), dataset sizes in number of tokens, and parameter counts without embedding layers from \newcite{KaplanMHB20}.}
    \label{fig:scaling_laws}
\end{figure}

Supporting Turing's observation, recent studies \cite{HestnessNAD17, KaplanMHB20} predict correlations between model size, dataset size, computational budgets, and the problem complexity \ac{NN}s can solve. \Cref{fig:scaling_laws} shows these three factors can accurately predict the test loss of transformer \ac{LM}s while the model shape and other hyperparameters are negligible.

However, in this work, we show how explicit incorporation of external knowledge from \ac{LKB} (\Cref{sec:chap3_method_wsd}) increases the performance of \ac{LM}s for downstream tasks while having less computational requirements and an equal dataset size and parameter count. Similar results for the \ac{WSD} task are present in the literature \cite{BlevinsZ20}.

The related fields of regularisation and pruning are concerned with the question of how to reduce model parameters without performance loss.
Pruning methods for \ac{LM}s show that specific attention heads in the transformer architecture are obsolete, and removing them yields marginal performance loss \cite{VoitaTMS19, MichelLN19}.
Supporting this hypothesis, \ac{KD} methods for \ac{LM}s \cite{SanhDCW19, WangWDY20} extract knowledge of large-scale models into significantly smaller ones while keeping most of their performance. 

To distill knowledge into smaller models requires exploring efficient architectures targeting expensive layers regarding the number of parameters and computational budgets. For transformer models, the most computationally expensive component is self-attention.
\newcite{Tay20} introduced a taxonomy to divide efficient architectures optimizing attention into five main categories:

\begin{description}
  \item[Fixed/Factorized/Random Parameter] methods reduce the self-attention matrix by limiting the field of view to fixed patterns, e.g., local windows and blocks.
  \item[Low Rank/Kernel] methods approximate the self-attention matrix by assuming a low rank to decompose it into a smaller dimension. Kernel methods rewrite self-attention to avoid computing large dot-products explicitly.
  \item[Recurrence] methods forward multiple sequences through the same model keeping previous states.
  \item[Memory] models are extended with side memory modules to store a temporary context for future processing.
  \item[Learnable Patterns] approximate the access pattern with data-driven approaches, e.g., clustering methods and hashing functions.
\end{description}

\noindent This work constructs a modified version of Longformer \cite{BeltagyPC20} which uses a combination of memory and fixed parameters. 
We choose Longformer because it performed superior over seven comparable \ac{LM}s in our related study \cite{Wahle21} and uses \ac{MLM}, making the model compatible with many transformer-based \ac{LM}s.
We also propose a lexical \ac{KD} technique that is general enough to be applied to many different \ac{LM}s (see \Cref{sec:chap3_method_wsd}).
\section{Knowledge Distillation}
In our first years as children, we learn language mainly from observing other humans and interacting with them. We expose ourselves to domain experts during education, whom we call teachers to learn from their knowledge and experiences. In large-scale \ac{LM}s the trend is different: novel models learn language patterns from scratch instead of leveraging the acquired knowledge of already trained ones.
We argue this course is inefficient and barely adds value to the semantic representations of a \ac{LM}. Many novel models train on very similar datasets (see \Cref{table:lm_litrev} in \Cref{sec:chap2_lm_kd}) with small performance gain in downstream tasks.

The field of \ac{KD} uses a similar teacher-student paradigm to the one employed by humans. When extracting acquired information (knowledge) into a smaller model (distillation), we call the process \acl{KD}. A crucial question in the domain of \ac{KD} is how to better represent knowledge in a \ac{NN}. In the following Section, we introduce essential knowledge concepts.

\subsection{Knowledge Concepts} \label{sec:chap2_knowledge_concepts}
We categorize knowledge concepts using three main categories related to our work:

\begin{description}
  \item[Logit-Based Knowledge] represents information encoded in the final predictions of a model, including semantic information.
  \item[Feature-Based Knowledge] exists in the hidden representations of a \ac{NN} that grow with abstraction towards the prediction layer.
  \item[External Knowledge] consists of structured and explicit sources created by humans to describe concepts.
\end{description}

\begin{figure} 
  \centering
  \includegraphics[width=.9\textwidth]{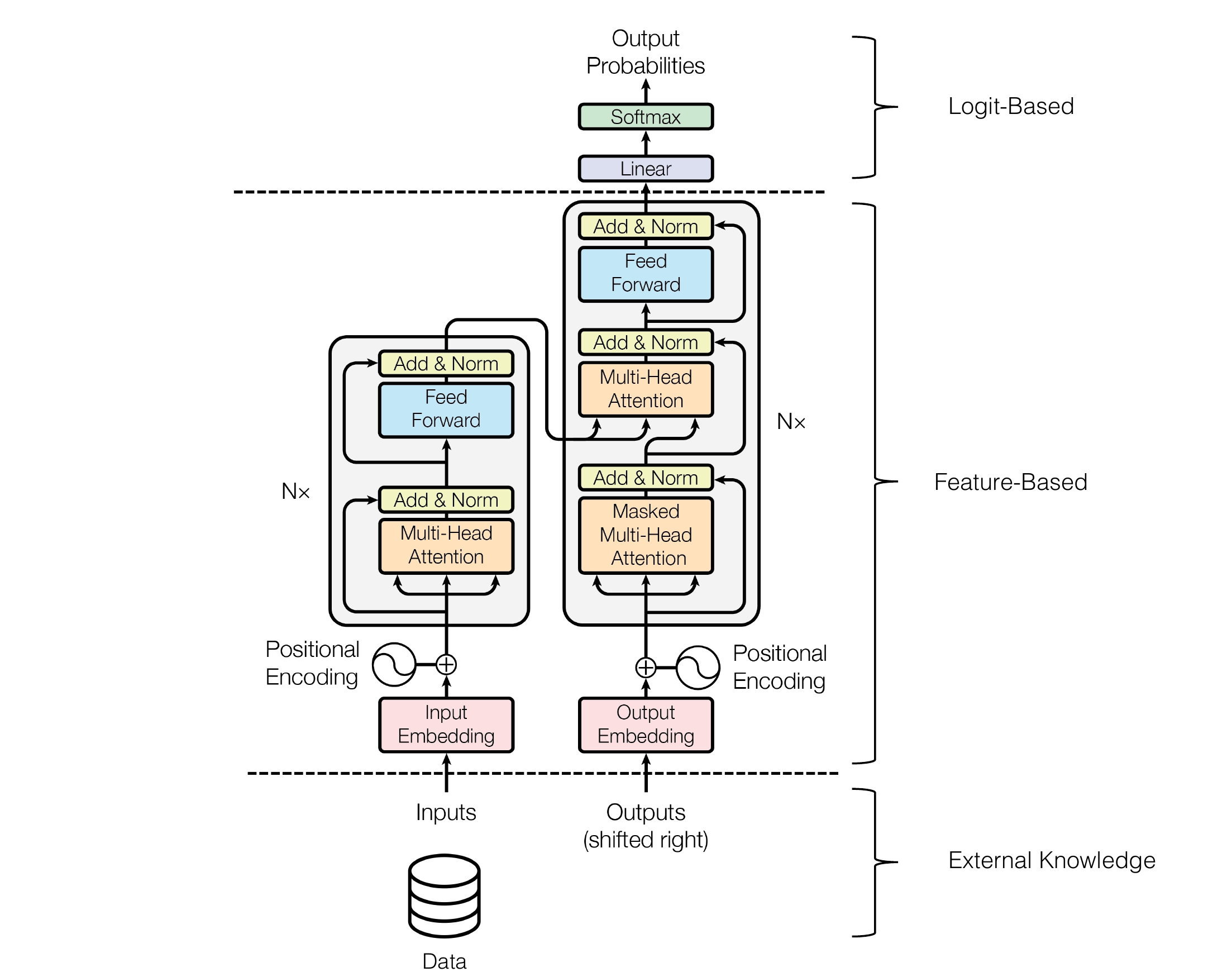}
  \caption[Knowledge concepts for the transformer architecture.]{Knowledge concepts visualized for the transformer architecture. This figure is an adaptation from Figure 1 in \cite{VaswaniSPU17}.}
  \label{fig:knowledge_concepts}
\end{figure}

\noindent The different types of knowledge for the transformer architecture are visualized in \Cref{fig:knowledge_concepts}. Other types of knowledge include relation-based \cite{YimJBK17a, LeeKS18} and self-distilled \cite{HahnC19, Zhang19} knowledge, which study the relation of features and data to improve representations within the same network.
We exclude these types of knowledge because they typically require the extracted network to be of the same size as the pre-trained network contradicting our primary research objective to optimize for a smaller model (see \Cref{sec:research_objective}). 

\subsubsection{Logit-Based Knowledge}
The information encoded in the probability distributions of a models' prediction layers represents logit-based knowledge. Logit-based \ac{KD} uses a soft distribution of a teacher's prediction layer as the supervision signal for a student \cite{HintonVD15}.
For example, in a mutually exclusive multi-class problem to classify whether a text describes one of five classes (e.g., Huskey, Eurasian, cat, car, bus), a trained teacher model assigns a high probability to the actual class (e.g., a Huskey), lower probabilities to similar animals (e.g., the Eurasian), lower but non-zero probabilities to other animals (e.g., cat), but close to zero probabilities to vehicles (e.g., car, bus). This semantic understanding \textit{dark knowledge} and is used to learn semantic features from the teacher model.

The logit-based method proposed by \cite{HintonVD15} mimics a softer version of a teacher model's predictions. \ac{NN}s typically learn classification using a softmax layer to produce probabilities $q_i$ from logits $z_i$ with a temperature parameter $T$ as \Cref{eq:softmax} shows.

\begin{equation} \label{eq:softmax}
    q_i = \frac{\exp(z_i/T)}{\sum_j \exp(z_i/T)}
\end{equation}

Usually, classification tasks set the temperature $T$ to 1. Larger values of $T$ produce a softer distribution (i.e., non-maximum logits increase and maximum logits decrease). Logit-based \ac{KD} uses the soft target distribution from the teacher network with $T>1$ to estimate a less strict output distribution. When the student model converged to model the teacher's soft distribution, $T$ is set back to 1. Additionally, logit-based methods regularly introduce the ground truth labels, proving to increase the student model's performance.

Logit-based \ac{KD} uses one part of the teacher network information, the final predictions. However, when working with \ac{NN}s, all hidden representations are available. Especially in deep learning, additional supervision signals using hidden representations improve training stability and reduce training time \cite{RomeroBKC15, JiaoYSJ20}.

\subsubsection{Feature-Based Knowledge}
Hidden representations in \ac{NN}s encode high-level features that increase in abstraction with layers closer to the final prediction \cite{BengioCV13}. Feature-based \ac{KD} uses these high-level features with regular supervision signals between hidden layers of the student and teacher model \cite{RomeroBKC15, ZagoruykoK17, KimPK18a}. 
Thin deep networks \cite{RomeroBKC15} propose hints, which are comparisons of hidden features from the student and teacher model using a regression loss. An extension of thin deep networks related to this work creates supervision signals using an attention score over features \cite{ZagoruykoK17}. The generalization of these attention scores calculates distributions of selectivity patterns and compares them rather than comparing features directly \cite{HuangW17}.
Probabilistic knowledge transfer explores the probability distribution of the data in the feature space instead of their sample representation \cite{PassalisT19}.

\subsubsection{External Knowledge}
Different from implicit knowledge found in trained \ac{NN}s, we can find knowledge in more explicit external sources, i.e., knowledge acquired and structured by humans to describe natural phenomena. For example, the Wikidata Knowledge Graph\footnote{\url{https://www.wikidata.org/}} contains 91 million entries describing relations of entities, e.g., humans, architectural structures, or chemical elements.
Knowledge-based methods for \ac{WSD} commonly explore the knowledge of \ac{LKB}. A popular example of a \ac{LKB} is WordNet \cite{Miller95, Fellbaum98a} which contains, e.g., \ac{POS} tags for words, word relations, and glosses\footnote{Gloss: A brief definition of a word sense.}.

External knowledge sources have two significant advantages over knowledge obtained through \ac{NN}s. First, they are interpretable by humans in a straight-forward manner. Second, their knowledge can be used explicitly to optimize learning a specific aspect of language (e.g., ambiguity). Unsupervised \ac{LM} pre-training proved to learn features of a language successfully, but with the cost of typically multiple billion tokens of optimization. With explicit knowledge from external sources, we show our models can gain superior semantic understanding while keeping the parameters low.

\subsection{Language Model Knowledge Distillation} \label{sec:chap2_lm_kd}
\ac{KD} has been extensively studied in the \ac{NLP} domain \cite{KimR16, HuPWH18}. As the usage of large-scale neural \ac{LM}s with many parameters increased, \ac{KD} gained much importance for the \ac{LM} research as well.

DistilBERT \cite{SanhDCW19} uses the logit-based \ac{KD} algorithm described in \cite{HintonVD15} to learn soft target probabilities of \ac{BERT} with the supervised \ac{MLM} loss. Additionally, DistilBERT optimizes its embedding output similarities with the cosine embedding loss yielding similar embeddings to the teacher model. To transfer prior knowledge of \ac{BERT} into a smaller model, DistilBERT keeps the weights of selected layers from the original model. As discussed, the hidden size has to remain equal between DistilBERT and \ac{BERT} when initializing \ac{BERT}'s weights. Therefore, DistilBERT controls model size through the number of layers.
The Patient \ac{KD} algorithm proposed by \newcite{SunCGL19} focuses on learning aggregate representations of the teacher model, i.e., the \texttt{[CLS]}-token in \ac{BERT}, rather than masked word soft-targets. The aggregate encodes the highest semantic information about a sequence and is typically used for classification tasks.

TinyBERT \cite{JiaoYSJ20} combines feature-based and logit-based approaches to imitate several larger layers into smaller ones (i.e., embedding layer, attention layer, hidden layer, prediction layer). In addition to pre-training, TinyBERT performs \ac{KD} at the fine-tuning step.
LadaBERT \cite{MaoWWZ20} reduces parameters and computational complexity with \ac{SVD} over the weight matrices and uses weight pruning \cite{HanPTD15, BlalockOFG20}. \newcite{MaoWWZ20}'s \ac{KD} methods are similar to TinyBERT using logit-based and feature-based knowledge.

MobileBERT \cite{SunYSL20} proposes bottleneck structures which are smaller alternatives to \ac{BERT}'s modules and balancing between multi-headed self-attention and feed-forward layers. To extract knowledge into the new structure, MobileBERT first trains a modified \ac{BERT} model, which uses the bottleneck architecture components. The student then distills knowledge from the modified \ac{BERT} layer-to-layer. Training the modified \ac{BERT} model adds additional training time, and performing layer-to-layer distillation fixes the number of layers and the teacher and student's hidden size.

MINILM \cite{WangWDY20} focuses on imitating the multi-headed self-attention modules of the teacher's transformer to gain feature-based knowledge. \newcite{WangWDY20} show high-level attention layers encode the highest semantic information. Thus, imitating layers close to the prediction step appears to yield similar performance with less supervision. Moreover, MINILM uses a teacher assistant \cite{MirzadehFLL20}, which stabilizes the \ac{KD} loss when the teacher model is much larger than the student model. 
MINILM has no requirement on the hidden size or the number of layers of the student and does not perform layer-to-layer distillation, making the student model more flexible.

\ac{BoT} \cite{XuLLL20} creates small substitutes for \ac{BERT}'s large layer modules. The model learns compact substitutes and large modules simultaneously in \ac{BERT} and increasingly replaces large modules with compact ones with advanced training progress. The smooth transition from large modules to compact modules results in an interaction of original modules and their compact replacements. This interaction proves to be superior to starting with many compact layers.

\ac{ERNIE} \cite{ZhangHLJ19} uses entities from the Wikidata Knowledge Graph with a token encoder and an entity encoder. The entity encoder uses an aggregation sub-layer to perform multi-headed self-attention over the tokens and entities, concatenating their resulting embeddings. The pre-training objective is a \ac{dEA}, which randomly masks token-entity alignments which the transformer needs to predict all corresponding entities based on aligned tokens.

The extension of \ac{ERNIE}, \ac{ERNIE} 2.0 \cite{SunWLF20} explores the incorporation of different pre-training objectives in a multi-task setup using explicit knowledge or self-supervised labels. Replacing training objectives typically reduces previously learned features as they are not needed to perform the new task. However, \ac{ERNIE} 2.0 maintains knowledge aspects by keeping objectives at a small percentage during training, resulting in semantically richer features.
We summarize the most popular presented transformer-based \ac{LM}s and provide a comprehensive overview in \Cref{table:lm_litrev}.
\begin{table}[htb]
\centering
\caption[Overview of transformer-based language models.]{Overview of transformer-based language models. Wikipedia refers to the official English Wikipedia Dump. The dash symbol "-" indicates that no official information was provided in the related research paper.}
\resizebox{\textwidth}{!}{
\begin{threeparttable}
    \begin{tabularx}{2\textwidth}{lXXXXX} \toprule

        \multirow{2}{*}{\textbf{Model}} & \multicolumn{3}{c}{\textbf{Dataset}} & \multicolumn{2}{c}{\textbf{Model}}\\
        \cmidrule(lr){2-4}
        \cmidrule(lr){5-6}

        {} & Corpora & Size / Tokens & Tokens / Epochs & Parameters & Training Procedure \\ \midrule

        \textbf{AlBERT} (\citeyear{LanCGG19}) & Books (\citeyear{ZhuKZS15}), Wikipedia & 13GB / 3.3b & 262b / 80 & large / xxlarge : 17m / 223m & \ac{MLM} + \ac{SOP} \\

        \textbf{BART} (\citeyear{LewisLGG19}) & Books (\citeyear{ZhuKZS15}), Wikipedia & 13GB / 3.3b & 2.2t (large) & base / large : 139m / 406m & Denoising Auto Encoder \\

        \textbf{BERT} (\citeyear{DevlinCLT19}) & Books (\citeyear{ZhuKZS15}), Wikipedia & 13GB / 3.3b & 137b / 40 & base / large : 110m / 340m & \ac{MLM} + \ac{NSP} \\
        
        \textbf{ELECTRA} (\citeyear{ClarkLLM20}) & Books (\citeyear{ZhuKZS15}), \ac{CC}\tnote{a}, ClueWeb 2012-B\tnote{b}, Gigaword 5 (\citeyear{NapolesGV12}), Wikipedia & 158GB / 33b & 419b / 12.7 & 335m & Replaced Token Detection \\

        \textbf{ERNIE} (\citeyear{ZhangHLJ19}) & Wikidata, Wikipedia & - / 4.5m & - / 1 & 114m & \ac{MLM} + \ac{NSP} + dEA \\

        \textbf{ERNIE 2.0} (\citeyear{SunWLF20}) & Books (\citeyear{ZhuKZS15}), Discovery Data (\citeyear{SileoVPM19}), Reddit, Wikipedia & 19GB / 7.9b & 78b /  10 & base / large : 110m / 340m & Knowledge Masking + Capital Prediction + Token Document Relation + Sentence Reordering \\
        
        \textbf{GPT-2} (\citeyear{RadfordWCL18}) & Open Web Text (\citeyear{RadfordWCL18}) & 40GB / 10.2b & - / - & medium / xl : 345m / 1.6b & Left-to-right \ac{LM} \\
        
        \textbf{Longformer} (\citeyear{BeltagyPC20}) & Books (\citeyear{ZhuKZS15}), \ac{CC}-Stories (\citeyear{TrinhL19}), Realnews (\citeyear{ZellersHRB19}), Wikipedia & 31.6GB / 6.4b & 39b / 6 & base / large : 149m / 435m & \ac{MLM} \\

        \textbf{MT-DNN} (\citeyear{LiuHCG19a}) & GLUE (\citeyear{WangSMH19}) & 4GB / 25m & 123m / 5 & base / large : 110m / 340m & Multi-Task Finetuning \\
        
        \textbf{RoBERTa} (\citeyear{LiuOGD19}) & Books (\citeyear{ZhuKZS15}), \ac{CC}-Stories (\citeyear{TrinhL19}), \ac{CC}-News\tnote{c} (Sept 2016 - Feb 2019), Open Web Text  (\citeyear{RadfordWCL18}), Wikipedia & 160GB / 33b & 2.2t / 66 & base / large : 125m / 360m & \ac{MLM} + large batches + full sentences + large \ac{BPE} vocab (50k pieces) \\
        
        \textbf{T5} (\citeyear{RaffelSRL20}) & C4 dataset (\citeyear{RaffelSRL20}) & 803GB / 165b & 34b / 0.2 & T5-large / T11 : 770m / 11b & fill-in-the-blank prediction \\

        \textbf{XLNet} (\citeyear{YangDYC19}) & Books (\citeyear{ZhuKZS15}), \ac{CC}\tnote{a}, ClueWeb 2012-B\tnote{b}, Gigaword 5 (\citeyear{NapolesGV12}), Wikipedia & 158GB / 32b & 2.2t / 66 & base / large : 110m / 340m & Permutational \ac{LM} \\
        
        \bottomrule 
    \end{tabularx}
    \begin{tablenotes}[para]
        \item[a] \url{http://commoncrawl.org/}
        \item[b] \url{https://lemurproject.org/clueweb12/}
        \item[c] \url{http://commoncrawl.org/2016/10/news-dataset-available/}
    \end{tablenotes}
\end{threeparttable}}
\label{table:lm_litrev}
\end{table}

This thesis shows the combination of logits and high-level features can extract knowledge into an efficient transformer architecture. Using external knowledge from \ac{LKB}, our model gains semantic understanding beneficial for many \ac{NLU} tasks.

\subsection{Multi-Teacher Knowledge Distillation} 
Different models provide individual knowledge aspects to perform tasks. The \textit{no free lunch theorem} states, there exists no single model that is best suited for all possible scenarios and data sets \cite{HoP02}. Instead, different \ac{LM} architectures are experts in specific language characteristics.

\newcite{HintonVD15} showed that ensembles of expert models are superior to a single generalized model. However, ensembles increase the cost of training and inference as they use multiple models simultaneously. To combine the knowledge of many experts, multi-teacher \ac{KD} combines the response from many teachers as the supervision signal (e.g., averaging the soft-targets \cite{HintonVD15}).

\newcite{TarvainenV17} average the teacher parameters instead of predictions and use a consistency loss between teacher's and student's predictions. The model keeps a moving average of label predictions on each training example and penalizes predictions inconsistent with the average.
\newcite{YouXXT17} extend learning soft-targets to features using a triplet loss by increasing the dissimilarity in feature representation between different examples and vice-versa.

In the domain of \ac{LM}s, some publications explore multi-task \ac{KD} \cite{LiuHCG19}, i.e., learning multiple tasks simultaneously to achieve more general representations, but ignore multi-teacher concepts. We see a gap in the literature to apply multi-teacher \ac{KD} to large-scale \ac{LM}s.

In this work, we transfer the concept of multi-teacher \ac{KD} to large-scale \ac{LM}s to gain their knowledge aspects. Our method weights the soft-target distribution of multiple teachers focusing on models with the highest prediction confidence.
\section{Applications}
To validate a \ac{LM} captures the characteristics of language, it needs to produce semantic representations beneficial for many \ac{NLU} tasks. \ac{NLU} benchmarks \cite{WangSMH19, SarlinDMR20} include a wide range of tasks and cover individual aspects of the target language, e.g., polysemy or text similarity.

In particular, \ac{WSD} is a difficult task to perform. \newcite{Navigli09} showed F1 scores above 80\% are difficult to surpass as the inter-annotator agreement (i.e., the percentage of words tagged with the same sense by two or more human annotators) is 67\% to 80\% for fine-grained, WordNet-style sense inventories. 
Polysemy frequently occurs in natural language, making it a crucial task to accomplish for any \ac{NLP} system. Furthermore, \ac{WSD} appears to be sensitive to changes in the \ac{LM} architecture and training scheme \cite{Wahle20} which makes it a suitable task to validate our proposed methods against different \ac{LM} architectures.

Another interesting application for \ac{LM}s lies in the domain of \ac{MPP} detection \cite{FoltynekRSM20, Wahle21}.
Academic plagiarism has become a pressing problem in our society. We know of more than 30 cases of academic plagiarism in Germany, including former minister of defense Karl-Theodor zu Guttenberg\footnote{\url{https://guttenplag.wikia.org/de/wiki/GuttenPlag_Wiki}}.
Modern plagiarists use machine-paraphrasing tools, which typically remain undetected by \ac{PD} systems using word-based or character-based text comparisons with an indexed corpus \cite{FoltynekMG19}. Real-world forms of plagiarism require detecting re-constructed and replaced parts of the text with probably unknown sources (e.g., Theses material).


\subsection{Word Sense Disambiguation} \label{sec:applications_wsd}

\ac{WSD} seeks to determine the meaning of words given a context and is a fundamental challenge in \ac{NLP} \cite{NavigliP12}. 
For example, the sentence ``I like java'' is ambiguous as ``java'' can refer to the programming language, an island, or coffee\footnote{according to the \ac{LKB} BabelNet 4.0 \url{https://babelnet.org/}}.

In knowledge-based methods \cite{CamachoColladosPN15}, \ac{LKB} (e.g., WordNet \cite{Fellbaum98a}, ConceptNet \cite{LiuS04}, BabelNet \cite{NavigliP12}) are used as a taxonomy to help categorize the relationship between words and their meaning. While unsupervised techniques \cite{ChaplotS18} do not rely on annotated data to perform disambiguation, supervised-based ones \cite{PasiniN20} explore human-labeled or automatically generated annotations. 

\ac{NN}-based models \cite{BengioDVJ03,MikolovSCC13,BojanowskiGJM17} have gained much attention in the \ac{NLP} community, mainly because of their success to capture latent semantic content and superior performance in tasks, such as word similarity \cite{NeelakantanSPM14}, text classification \cite{RuasFGd20}, and topic categorization \cite{PilehvarCNC17}. \newcite{RuasGA19} combine WordNet \cite{Fellbaum98a} and \ac{W2V} \cite{MikolovSCC13} to represent word senses in fixed-sized length vectors. Their \ac{MSSA} algorithm combines the vector representation for each word in WordNet's glosses using a general pre-trained word embeddings model. \ac{MSSA} averages the word representations in a context sliding window to select the meaning of a word with the highest similarity within its adjacent neighbors. However, the technique does not explore the benefits of transfer learning between different tasks. Likewise, this thesis considers the glosses in WordNet but relies on the \ac{LM} do identify the correct sense of a word given its context with an end-to-end approach.

Using a large-scale \ac{LM}, GlossBERT (with their best performing method Sent-CLS-WS) \cite{HuangSQH19} uses WordNet's glosses to fine-tune \ac{BERT} for the \ac{WSD} task. GlossBERT classifies a marked word in a sentence into one of its possible definitions.
For each word sense of an ambiguous word, GlossBERT creates a pair of the context and a gloss using the \ac{BERT} tokenizer. Their method inserts two supervision signals: first, highlighting ambiguous tokens with two unique tokens, and second, repeating the ambiguous tokens with the gloss definition.
\newcite{DuQS19} fine-tune \ac{BERT} similarly with the encoder and a classifier but without additional supervision.

KnowBERT \cite{PetersNLS19} incorporates \ac{LKB} into \ac{BERT} (e.g., WordNet) with a \ac{KAR} mechanism. The \ac{KAR} component uses mention-spans to retrieve entity embeddings from a \ac{LKB}, update the mention-span embeddings with the linked information and recontextualize the entity embeddings with altered multi-head attention. Their multi-head self-attention is modified so a word-piece can attend to all entity embeddings in the context. \newcite{PetersNLS19} best-performing model, i.e., KnowBERT combining Wikipedia and WordNet (KnowBERT-W+W), achieves better results than both \ac{BERT} models ($base$ and $large$). However, KnowBERT-W+W adds over 400 million parameters compared to \ac{BERT}$_{base}$ and is 32\% slower. Other approaches combining \ac{BERT} and WordNet for the \ac{WSD} task, but less related to our contributions, are also discussed in the literature \cite{VialLS19}.

\newcite{BlevinsZ20} propose \ac{BEM}, a model to encode a target word with its surrounding context and the gloss of each sense in separate encoders. Their two encoders are learned simultaneously from the \ac{WSD} objective. \ac{EWISER} \cite{BevilacquaN20}, an extension of \cite{KumarJST19} uses different word embeddings as input to include semantic relations (e.g., hypernym) in their structure. \ac{LMMS} \cite{LoureiroJ19} combines $k-$NN using the \ac{MFS} in WordNet with \ac{BERT} embeddings. \ac{GLU} \cite{HadiwinotoNG19} integrates contextualized word representations in \ac{WSD}. The previous techniques enhance semantic representations via context, external knowledge, or \ac{MFS}, but do not explore generalization to other \ac{NLP} tasks. 

The methods proposed in this thesis do not require recurrent embeddings adjustments from the \ac{LKB} nor use word-piece attention, resulting in less overhead to the system. 
We show by choosing the best-suited model for \ac{WSD} and adjusting the training procedure, our methods outperform preceding techniques in \ac{WSD} and obtains the highest score in 7 out of 9 \ac{NLU} tasks.

\subsection{Machine-Paraphrased Plagiarism Detection} \label{sec:applications_mpp}

Plagiarism is a severe form of academic misconduct and a pressing problem for educational and research institutions, publishers, and funding agencies \cite{FoltynekMG19}. 
To counteract plagiarism, many institutions employ \ac{PD} systems. These tools reliably identify duplicated text yet are significantly less effective for paraphrases, translations, and other concealed forms of plagiarism \cite{MeuschkeG13, FoltynekDAR20}. 

Studies \cite{RogersonM17, PrenticeK18} show an alarming proportion of students employs online paraphrasing tools (often referred to as text spinning tools) to disguise text taken from other sources. These tools are expected to employ \ac{NN} approaches to change a text, e.g., replacing words with potential synonyms \cite{ZhangWV14}. \newcite{ZhangWV14} presented a tool intended to perform Search Engine Optimization \cite{MaderaGM14} by inflating a website's PageRank. The idea is to use the promoted page's content to create bogus links to an advertised website. Paraphrasing tools serve to alter the content, such that search engines do not recognize the fraudulent websites as duplicates and consider them to calculate the PageRank of the promoted site.

In academia, paraphrasing tools help to mask plagiarism, facilitate collusion, and help ghostwriters with producing work that appears original. Paraphrasing tools severely threaten text-matching software effectiveness, which is a crucial support tool for ensuring academic integrity. The academic integrity community calls for technical solutions to identify the machine-paraphrased text as one measure to counteract paraphrasing tools \cite{RogersonM17}.
The International Journal for Educational Integrity recently devoted a special issue\footnote{\url{https://edintegrity.biomedcentral.com/mbp}} to this topic. 

To contribute to the solution for academic integrity, we devise an automated approach that reliably distinguishes human-written from machine-paraphrased text and provide the solution as a free and open-source web application. A recent short paper \cite{FoltynekRSM20} reports on our fellow researchers' preliminary experiments using one paraphrasing tool and a Wikipedia corpus. In this thesis, we analyze two new collections created from research papers on arXiv\footnote{\url{https://arxiv.org}} and graduation Theses of \textit{English as a second language} students. Additionally, we explore a second paraphrasing tool for generating obfuscated samples and eight neural language models based on the transformer architecture in all datasets. The recent neural language models surpass all preliminary \ac{ML} techniques in every test set. Compared with two plagiarism detections systems, Turnitin\footnote{\url{https://www.turnitin.com/}}, which has the largest market share, and PlagScan\footnote{\url{https://www.plagscan.com/}} which is one of the best-performing systems \cite{FoltynekDAR20}, neural language models detect short paragraphs of machine-paraphrased text with higher confidence. 
\chapter{Methodology} \label{Chapter3}

The main objective of this thesis is to explore \ac{KD} using different knowledge sources. This chapter presents a method to distill knowledge from large-scale \ac{LM}s (\Cref{sec:multi_teacher}) into an efficient model and two methods to distill knowledge from lexical databases (\Cref{sec:chap3_method_wsd}).
We introduce a general framework to perform \ac{WSD} with a variety of \ac{LM}s relying on the transformer architecture and show how their improved semantic understanding can be applied to other \ac{NLU} tasks (\Cref{sec:chap3_method_wsd}).
Finally, we present two new data collections to evaluate \ac{MPP} detection and explore two paraphrasing tools for generating obfuscated samples. We extend the application of \ac{ML} techniques using word embeddings and propose a system to detect \ac{MPP} with transformer \ac{LM}s (\Cref{sec:chap3_method_mpp}).

\section{Knowledge Distillation with Multiple Language Models} \label{sec:multi_teacher}

\begin{figure}[!hbt] 
    \centering
    \includegraphics[width=\textwidth]{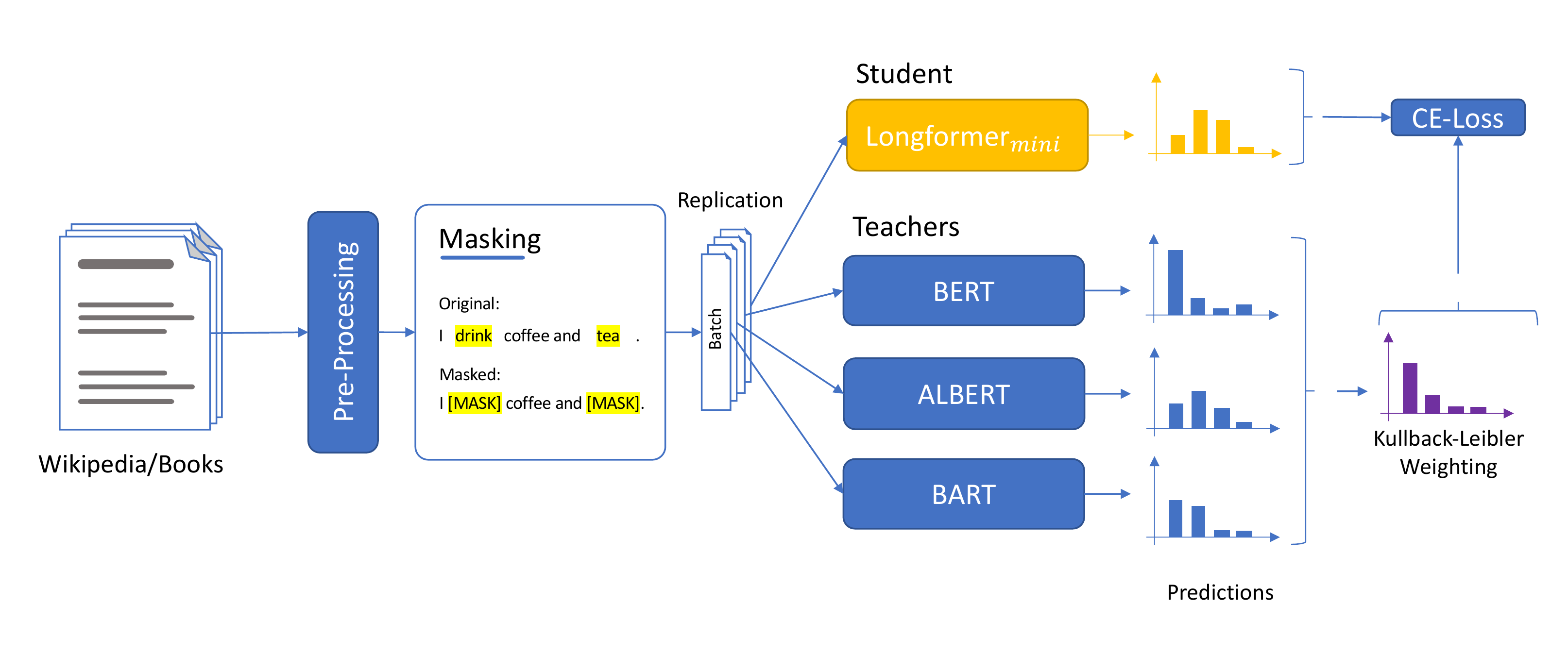}
    \caption[Overview of the multi-teacher Knowledge Distillation method.]{Overview of the multi-teacher \ac{KD} method.}
    \label{fig:kd_multi_teacher}
\end{figure}

\ac{KD} mainly explores the logits and features of a single \ac{LM} with varying student model sizes (see \Cref{sec:chap2_lm_kd}). This work aims to combine the individual knowledge of multiple teacher models with a new prediction weighting approach. The goal of our \ac{KD} technique is to provide a method for training a small model that incorporates the knowledge of multiple large models and thus, can be used by the community as a replacement for its expensive counterparts. We provide an overview of our method in \Cref{fig:kd_multi_teacher}.

\subsection{Student Architecture and Initialization}
Multi-teacher \ac{KD} requires the student model's training to be compatible with many teachers' training approaches. \ac{LM}s often use pre-training objectives similar to \ac{MLM} (see \Cref{table:lm_litrev}) and output a probability distribution over a standardized sub-word vocabulary for each corrupted token. Furthermore, the student model is smaller than its teachers and needs to use its parameters efficiently.

We initialize the student with the $base$ configuration of Longformer \cite{BeltagyPC20} as the training architecture is compatible with many other transformers (e.g., \ac{BERT}, \ac{ALBERT}, \ac{RoBERTa}), while using an efficient method to calculate attention.
Furthermore, Longformer performed superior over other architectures in our related experiments \cite{Wahle21},.

To control the student's trainable parameters, we reduce the number of layers to 6 and retain all remaining configurations of the $base$ model. Following \newcite{SanhDCW19}, we decrease layer count rather than model width or attention head size to initialize pre-trained layers from the $base$ model. We keep the prediction layer of Longformer$_{base}$ and remove every second layer towards the network's input while keeping all weight matrices. We name the resulting model Longformer$_{mini}$.

\subsection{Distillation Method}
The student network gains knowledge about the target language in two ways. First, it mimics the soft probability distribution of masked tokens using a custom weighted average. Second, we use the last high-level hidden features of each teacher as studies \cite{WangWDY20,JiaoYSJ20} proved it to be superior over single logit feedbacks.

We propose a logit-based supervision signal weighting the probability of each teacher network by confidence. Teachers with a high probability for the correct class receive a larger proportion in the averaged target distribution, while low probabilities lead to a smaller proportion.
We use the \ac{KL} divergence \cite{KullbackL51} to compare the classification distribution vector of the $i$-th teacher model $\hat{y}_i$ and the groud truth label vector $y$ to weight confident and unconfident predictors with high and low proportions in the target distribution respectively (see \Cref{eq:weighted_teacher_supervision}).

\begin{align} \label{eq:weighted_teacher_supervision}
    y_{target} &= \sum_{i=0}^n KL(\hat{y}_i || y) \hat{y}_i \\
    &= \sum_{i=1}^n \hat{y}_{i} \sum_{j=1}^m \log\left(\frac{\hat{y}_{i,j}}{y_j}\right)
\end{align}

The student uses the weighted prediction distribution $y_{target}$ together with regular introductions of the ground truth label $y$ as targets for the students \ac{LM} objective \cite{HintonVD15}. The model learns features of one layer prior to the prediction using the \ac{MSE} loss \cite{MSE}. \Cref{algorithm:kd} details the procedure, where the \textit{forward} function obtains the predictions of a model with given inputs, and \textit{maskfunction} replaces words with mask tokens at a certain probability (see \Cref{sec:chap3_kd_setup} for hyperparameter details).

\begin{algorithm}[htb]

    \KwData{Training Data $D \sim \mathcal{X} \times \mathcal{Y}, \mathcal{X} \subset \mathbb{N}$ represents the input words as indexes of the vocabulary $V$. $\mathcal{Y} \subset \mathbb{N}$ represents the target masked words as indexes of the vocabulary $V$.}
    \KwIn{Teachers $T=(t_1,...t_n)$. Student $s$. Ground truth update step $l$.}
    \For{$(x,y) \in D$; $k=1$ \KwTo $|D|$}{

        $\widetilde{x} \leftarrow maskfunction(x)$ \\
        $\hat{y}_{student} = forward(s, \widetilde{x})$ \\

        \For{$t_i \in T$}{
            $\hat{y}_i \leftarrow forward(t_i, \widetilde{x})$ \\
            $y_{target} \leftarrow y_{target} + \hat{y_i} \sum_{i=0}^m \log\left(\frac{\hat{y}_{i,j}}{y_j}\right)$
        }
        \If{$k \mod l = 0$}{
            $l \leftarrow $CE$(\hat{y}_{student}, y)$ \\
        }
        \Else{
            $l \leftarrow $CE$(\hat{y}_{student}, y_{target})$ + MSE($\hat{h}_{student}, h_{target}$) \\
        }
        backward($l$) \\
        optimizer.step() \\
    }
    \caption{{\bf Multi-Teacher \ac{KD} algorithm for \ac{LM}s} \label{algorithm:kd}}

\end{algorithm}

As our modified version of the Longformer model is smaller than large-scale \ac{LM} teacher candidates and uses a local windowed and global attention scheme that is different from self-attention, we add no supervision signal for attention layers.
Although \newcite{WangWDY20} showed their feature approximation method overcomes changes in model size by using a teacher assistant, the feature approximator for our method must not only learn a function to map multiple attention layers into a single one but also map self-attention to local and global attention.
We presume the approximated target function of attention scores would include larger amounts of noise than distilling the same attention structure, impacting the training convergence.

Logit-based and feature-based knowledge in the final layers are not bound to the attention scheme or model size used and do not require a feature approximator.
Furthermore, our presented method appears more accessible, as the interpretation of prediction probability distributions is intuitive.

\subsection{Training and Testing Datasets}
Transformer models typically perform training using a combination of Wikipedia articles and books as \Cref{table:lm_litrev} emphasizes.
Although recent studies show an increase in dataset size (from $\approx$10 times \cite{LiuOGD19} up to $\approx$80 times \cite{RaffelSRL20}) and model size improves performance, we consider the small gain on average as ineffective. Instead, we construct a compact dataset composed of the official Wikipedia dump from October 2020\footnote{\url{https://en.wikipedia.org/wiki/Wikipedia:Database_download}} and the Books Corpus \cite{ZhuKZS15} as a cheaper alternative covering long and short documents from different domains. The English Wikipedia contains $\approx$6.2m articles with $\approx$3.7 billion words and an average of about 600 words per article. The Books Corpus contains $\approx$11k books with $\approx$1b words, with $\approx$90k words per book on average.
We pre-process all documents with the official \ac{BERT} tokenizer ync split 20\% of documents from Wikipedia and books at random as the validation dataset. We provide more details about the full benchmark in \Cref{table:kd_dataset}.

\begin{table}[htb]
    \caption[Overview of the multi-teacher Knowledge Distillation training corpus.]{Overview of the multi-teacher \ac{KD} training corpus.}
    \centering
    \resizebox{\textwidth}{!}{
        \begin{tabular}{lrrrrrr} \toprule
        \textbf{Dataset} & \textbf{Documents} & \textbf{Words} & \textbf{Words per Doc.} & \textbf{Vocab.} & \textbf{Mean Words per Sent.} \\ \midrule
        Wikipedia (Oct. 2020)  & 6.16m & 3.69b & 599 & 550.01k & 11 \\
        Books Corpus (\citeyear{ZhuKZS15}) & 11.04k & 984.85m & 89.22k & 1.31m & 13 \\ \bottomrule 
    \end{tabular}}
    \label{table:kd_dataset}
\end{table}


\subsection{Setup} \label{sec:chap3_kd_setup}
Our rationale for choosing teachers was to explore models that cover different training architectures but are compatible with our students' pre-training objectives. Student compatibility requires the teacher to a output probability distribution for corrupted tokens and to use the same vocabulary.
We choose \ac{BERT} \cite{DevlinCLT19} as a well-researched baseline model, \ac{ALBERT} \cite{LanCGG19} because of its increased training capacity using shared parameters, and BART \cite{LewisLGG19} due to its different denoising training approach combined with \ac{AR} language modeling. Furthermore, the three teacher models trained on similar datasets to ours (see \Cref{table:lm_litrev}), which we assume increases the probability of creating meaningful predictions than models exploring other domains, not including Wikipedia or Books. 

We applied best practices for model pre-training proposed by \newcite{LiuOGD19}, using a large batch size of 256 examples with gradient accumulation and dynamic masking. The remaining configuration details are as follows: a sequence length of 512 tokens, the AdamW optimizer with a learning rate of $2\times 10^{-5}, \beta_1 = 0.9, \beta_2 = 0.999, \epsilon = 1e-8$, the temperature $T=2.5$, a masking probability of $0.15$, a ground truth update step of $l=100$, and PyTorch's native automated mixed-precision format. The experiments use 8 NVIDIA Tesla V100 \ac{GPU}s with 16GB memory per card.

\subsection{Implementation Details}
Emphasizing the problem of model size, a single base-sized \ac{LM} (e.g., \ac{BERT}$_{base}$, $L=12, H=768, A=12$) together with the student model barely fits into the \ac{GPU} memory of a high-end NVIDIA Tesla V100 GPU with 16GB memory and mixed half-precision. When considering multiple teacher models and potentially larger models for future experiments, we need to construct a synchronous parallel setup for training in two ways. 

\subsubsection{Model Parallelism}
Instead of processing teacher and student models on the same \ac{GPU}, our method dedicates a single processing unit to each model. At each training step, we replicate the input sample once per unit, and each model processes a forward pass. We gather the final predictions in a shared memory pool and process the loss calculation, backward pass, and update rule on the student's side as it uses less memory and computation resources due to its optimized attention scheme and small model size.

\subsubsection{Data Parallelism}
To further scale training to multiple \ac{GPU}s yielding larger batch sizes with improved training time and convergence \cite{LiuOGD19}, we parallelize the model set up by replicating each model on a second \ac{GPU} and average their gradients in each step. To decrease training time, we pinned the unallocated memory of each \ac{GPU} and loaded new training samples asynchronously while the \ac{GPU} processed the current sample.
\section{Incorporating Lexical Knowledge into language Models} \label{sec:chap3_method_wsd}

\begin{figure}[!hbt] 
    \centering
    \includegraphics[width=.65\textwidth]{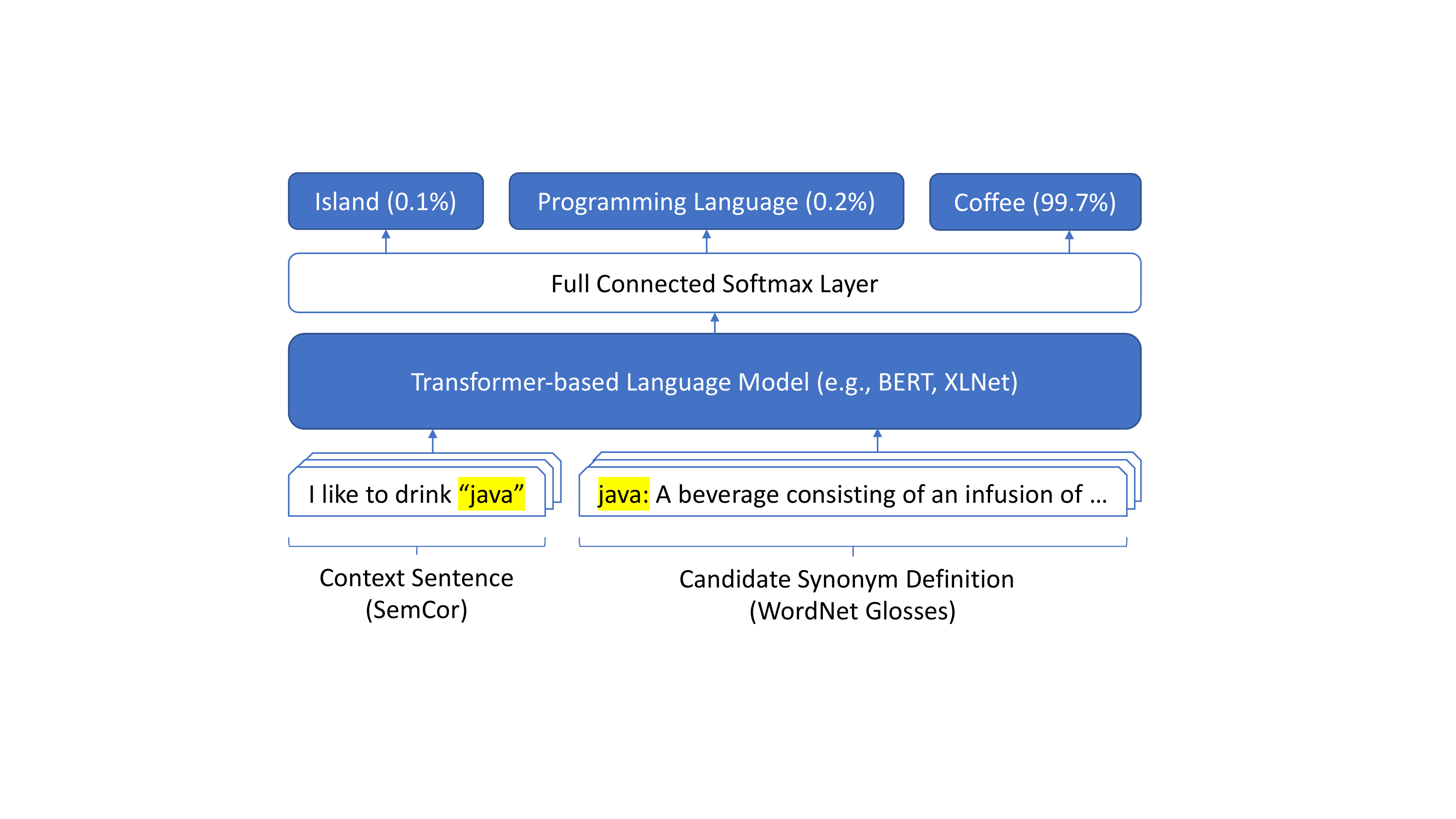}
    \caption[Overview of the Word Sense Disambiguation method.]{Overview of the \ac{WSD} method.}
    \label{fig:kd_lexical}
\end{figure}

Humans use many explicit sources to gain knowledge about concepts, e.g., a lexicon describes different natural phenomena in an organized way. Knowledge-based methods for \ac{WSD} use semantic information to determine word senses.
In the following, we present a method using the knowledge about word senses in \ac{LKB} to improve the semantic understanding of \ac{LM}s. 
Our method addresses research \ref{itm:researchtask2}, increasing semantic understanding and boosting task performance (see \Cref{sec:chap4_eval_wsd}). We provide an overview of the method using the lexical database WordNet \cite{Miller95,Fellbaum98} and the annotated SemCor dataset \cite{MillerLTB93} in \Cref{fig:kd_lexical}.

Current state-of-the-art methods in \ac{WSD} focus primarily on the \ac{WSD} domain without using the knowledge about ambiguous words for other \ac{NLP} tasks (see \Cref{sec:applications_wsd}). We suggest incorporating two objectives into the training of \ac{WSD} to maintain \ac{LM} capabilities while learning the disambiguation of words. Thus, the model obtains superior representations that benefit other tasks and domains as most benchmarks include a high fraction of polysemous words (see \Cref{AppendixA}).  

Our literature analysis showed related works in the domain of \ac{WSD} using neural \ac{LM}s are typically based on \ac{BERT} (see \Cref{sec:applications_wsd}). \ac{BERT} is a strong baseline, but recent studies show the model has not reached its full capacity; its training scheme still offers opportunities for improvement \cite{LiuOGD19}.
We introduce a method to perform \ac{WSD} with arbitrary \ac{LM}s and explore architectural changes to increase our model's performance (\Cref{sec:lmgc}). 

Furthermore, related \ac{WSD} methods \cite{HuangSQH19} use a sequential binary prediction head which requires $n$ forward passes with the model for one ambiguous word (with $n$ candidate word senses). With models containing multiple hundred million parameters, sequential processing becomes a bottleneck.  We form the prediction step from sequential binary classification to parallel multi-classification to construct a more natural prediction head as previous literature showed \cite{KagebackS16}.

\subsection{Training and Testing Datasets} \label{sec:wsd_datasets}
We use the SemCor 3.0 \cite{MillerLTB93} dataset as the training corpus for all \ac{WSD} experiments. SemCor 3.0 is one of the largest manually annotated datasets with approximately 226k word sense annotations from WordNet \cite{Miller95} for all open-class parts-of-speech. Each sentence in SemCor has multiple word annotations, and each word annotation consists of one or more sense annotations. The SemCor 3.0 corpus is well studied in the \ac{WSD} literature \cite{HuangSQH19, PetersNLS19}. 
We validated \ac{LM}s trained on SemCor 3.0 with\newcite{RaganatoCN17}'s evaluation framework, a set of five standardized test sets: Senseval-2 \cite{EdmondsC01}, Senseval-3 \cite{SnyderP04}, SemEval-2007 \cite{StrapparavaM07}, SemEval-2013 \cite{NavigliJV13}, and SemEval-2015 \cite{MoroN15}. We provide a detailed overview of these datasets in Table \ref{table:dataset_sem}.

\begin{table}[htb]
\caption[Overview of the SemCor training corpus.]{SemCor training corpus details: general statistics (left) and class distribution for binary classification of word senses (right).}
\centering
\resizebox{.8\textwidth}{!}{
    \begin{tabular}{lrrrrrrrrr} \toprule
    \multirow{2}{*}{\textbf{Dataset}} & \multicolumn{5}{c}{\textbf{POS Tags}} & \multicolumn{2}{c}{\textbf{Class dist.}} \\
    \cmidrule(lr){2-6}
    \cmidrule(lr){7-8}
    {} & Noun   & Verb   & Adj. & Adv. & Total & Pos. & Neg. \\ \midrule
    SemCor  & 87k & 88.3k & 31.7k   & 18.9k & 226k & 226.5k & 1.79m \\
    SE2     & 1k & 517 & 445 & 254 & 2.3k & 2.4k & 14.2k \\
    SE3     & 900 & 588 & 350 & 12 & 1.8k & 1.8k & 15.3k \\
    SE7    & 159 & 296 & 0 & 0  & 455 & 459 & 4.5k \\
    SE13    & 1.6k & 0 & 0 & 0 & 1.6k & 1.6k & 9.7k \\
    SE15    & 531     & 251     & 160       & 80 & 1k & 1.2k & 6.5k \\        \bottomrule 
\end{tabular}}
\label{table:dataset_sem}
\end{table}

To validate our proposed methods achieve high performance in \ac{WSD} while using the acquired knowledge for other \ac{NLP} tasks, we used the \ac{GLUE} benchmark. \ac{GLUE} \cite{WangSMH19} is a collection of eight language understanding tasks widely used in the language modeling domain \cite{DevlinCLT19, LanCGG19, LiuOGD19} to validate transfer learning capabilities of language models.
All \ac{GLUE} tasks are single sentence or sentence pair classification, except STS-B, which is a regression task. All classification tasks are binary classification except for MNLI, which has three classes.

\subsection{Language Model Gloss Classifiction (LMGC)} \label{sec:lmgc}

With \ac{LMGC}, we propose a model-independent end-to-end \ac{WSD} approach to classifying ambiguous words from sentences into one of WordNet's glosses. This approach enables applying different \ac{LM}s for \ac{WSD}. \ac{LMGC} addresses a problem resulting from imbalanced examples by using the focal loss function \cite{LinGGH17}, a state-of-the-art method to avoid accumulated gradients from negative examples. By choosing the most suitable model for \ac{WSD} from eight different \ac{LM}s, we show superior performance over \ac{BERT} (\Cref{sec:lmgcm}).

In the \ac{LMGC} task we pair (i) a sentence containing a polysemous target word with (ii) a prospective gloss definition from a lexical database (e.g., WordNet) for this word. Let $W = (w_1, \dots, w_n)$ be a word sequence of a sentence, where $w_i$ is a polysemous word. WordNet provides $S_i = \{s_1, \dots, s_m\}$ possible senses for word $w_i$ with corresponding glosses $G_i = \{g_1, \dots, g_m\}$, which are also individual sentences.

We build an annotated corpus using WordNet glosses to perform \ac{LMGC}.
For a sentence $W$, each polysemous word $w_i$ points to correct WordNet senses $\widetilde{S}_i \subset S_i$. We retrieve the gloss candidates $G_i$ for each synset of the word to create $|G_i|$ pairs of the sentence $W$ and glosses $g_j$, with $g_j \in G_i$. If $g_j$ corresponds to the desired meaning of our polysemous target word $w_i$ (which is the case if $s_j \in \widetilde{S}_i$), we classify the pair as a positive example, otherwise as negative. As the number of correct senses for a polysemous word is much smaller than the total number of senses, labels are often imbalanced (see \Cref{sec:wsd_datasets}).

The input sequence used in \ac{LMGC} follows the same configuration as in its underlying transformer. Each input sequence starts with the aggregate token (e.g., \texttt{[CLS]} for \ac{BERT}), followed by a sentence and a gloss definition. The sentence and gloss are separated with a unique separator token and tokenized using WordPiece \cite{SchusterN12}.

To perform the classification, \newcite{DuQS19} used the token embeddings of polysemous words. However, \newcite{HuangSQH19} showed the classification on top of the aggregate representation improves the results in \ac{WSD}.
Therefore, we obtain the final hidden representation of the aggregate token, namely $C \in \mathbb{R}^H$; where $H$ is the embedding size. 
We apply a weight matrix $W_{{\scaleto{LMGC}{4pt}}} \in \mathbb{R}^{H \times 2}$ transforming $C$ (together with a bias $B_{{\scaleto{LMGC}{4pt}}} \in \mathbb{R}^{2}$) into a binary space and calculate the probability to whether the gloss is appropriate for the ambiguity in context as \Cref{eq:binary_probs} shows.

\begin{equation} \label{eq:binary_probs}
    p = \textrm{softmax}(C~W_{\scaleto{BGP}{4pt}}^T + B_{\scaleto{BGP}{4pt}})
\end{equation}

To optimize the model for learning whether a gloss is correct for the annotated ambiguous word, \newcite{HuangSQH19} and \newcite{DuQS19} used the standard cross-entropy loss function (\Cref{eq:cross_entropy}).

\begin{align} \label{eq:cross_entropy}
    \textrm{CE}(p,y) = \begin{cases}
        - \log(p)& \textrm{if y = 1}\\
        1 - \log(1-p)& \textrm{otherwise}
    \end{cases}
\end{align}

The class imbalance resulting from correct word senses in relation to candidates hinder the cross-entropy loss during training (see \Cref{table:dataset_sem}). Easily classified negatives comprise parts of the loss, which obstructs the gradient direction. Thus, we propose to apply a weighted loss function, a popular approach to mitigate the class imbalance in object detection from camera images \cite{RenHGS16,RedmonDGF16}. The focal loss \cite{LinGGH17} (\Cref{eq:focal_loss}) reshapes the cross-entropy loss function, giving less weight to easy examples and forces the training to focus on challenging polysemous words.

\begin{equation} \label{eq:focal_loss}
    \text{FL}(p, y) = \begin{cases}
        -(1-p)^{\gamma} \log(p)& \textrm{if y = 1}\\
        -(1+p)^{\gamma} \log(1 - p)& \textrm{otherwise}
    \end{cases}
\end{equation}

We add additional supervision to the input sequence according to \newcite{HuangSQH19} with two signals: (1) highlighting the ambiguous tokens with two special tokens and (2) adding the polysemous word before the gloss.

\subsection{Language Model Gloss Classification with MLM (LMGC-M)} \label{sec:lmgcm}

\begin{figure}[tb] 
    \centering
    \includegraphics[scale=0.8]{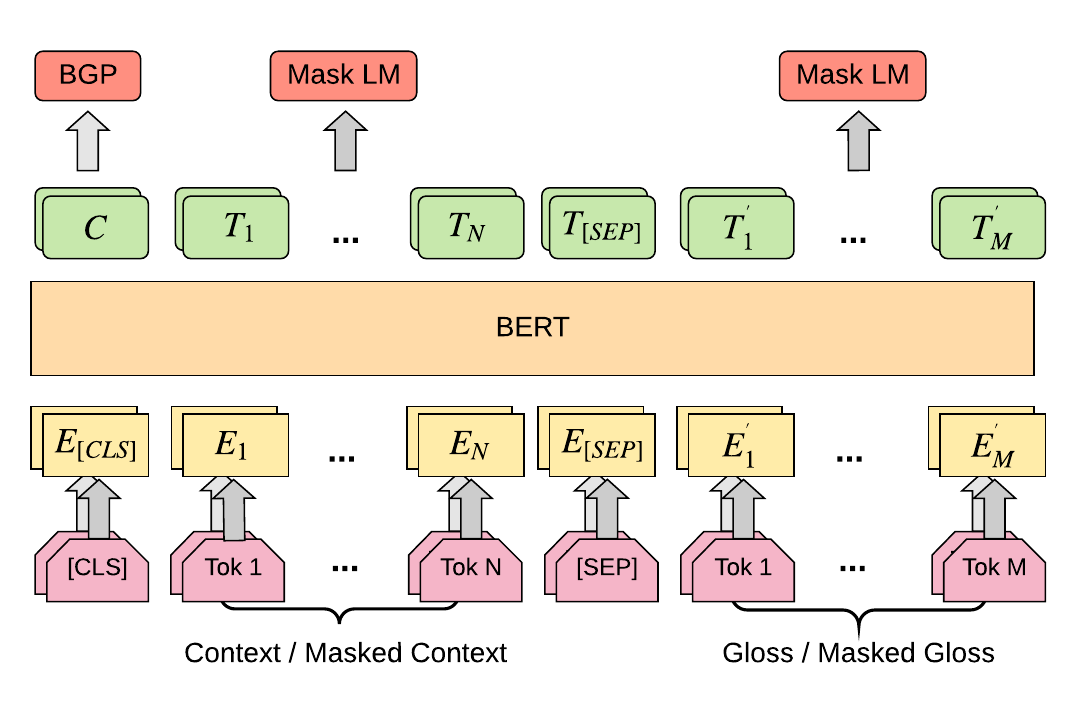}
\caption[Language Model Gloss Classification with Masked Language Modeling.]{\ac{LMGC-M} forwards context-gloss pairs and its masked versions through a \ac{LM} (e.g., \ac{BERT}) to perform a weakly supervised binary classification \ac{LMGC} and \ac{MLM}.}
\label{fig:methods}
\end{figure}

Pre-trained models serve as a valuable feature extractor that can be optimized to new task-specific objectives in a straight-forward manner \cite{RadfordWCL18, DevlinCLT19}. However, optimizing a specific task disregard previously learned \ac{LM} capabilities (see \Cref{sec:chap4_eval_wsd}). The \ac{WSD} task is special, as we can assume when a \ac{LM} is capable of disambiguation, its ability to perform well on other tasks may increase. For this reason, we aim to keep language modeling features while learning \ac{WSD}. Transfer learning between language modeling and \ac{WSD} increases the likelihood of understanding polysemous words in other related tasks, improving the overall performance. 
In \Cref{AppendixA}, we discuss the correlation between \ac{WSD} and other \ac{NLU} tasks concerning the number of polysemous words.

The task-specific objective of \ac{LMGC} is to predict whether the appended gloss explains the correct meaning of a word in a sentence. The corresponding objective function is the focal loss function between the network's prediction and the ground truth label as \Cref{eq:focal_loss} shows. In \ac{LMGC-M}, we perform a parallel forward pass through transformer \ac{LM} with similar sentence-gloss pairs. The first pass uses the original pair of a context sentence and gloss and applies a weight matrix to the final aggregate embedding resulting in binary classification. In the second pass, we replace words in the sentence and gloss with a mask token, a random token, or the same token with equal probabilities with the same configuration as \newcite{DevlinCLT19}. 
The \ac{LMGC-M} procedure is exemplified in \Cref{fig:methods}.

We sum the two objectives, i.e. the task-specific \ac{WSD} focal loss and the \ac{MLM} cross-entropy loss, to obtain the final loss function $L_1$ described in \Cref{eq:focal_MLM_loss}

\begin{equation} \label{eq:focal_MLM_loss}
    L_1(\hat{y}, y) = \text{FL}(\hat{y},y) + \sum_{k=1}^m \text{CE}(\hat{y}^{(mask)}_k,{y}^{(mask)}_k)
\end{equation}

where $m$ is the number of masked words, $\widetilde{y}_j \in \{0, 1\}^V$ is the one-hot encoding for a masked word $w_j$, and $\widetilde{p}_j \in \mathbb{R}^V$ is the probability vector for $w_j$.

\subsection{Setup}
We initialized all models using the base configuration of its underlying transformer (e.g., XLNet$_{base}$, $L=12, H=768, A=12$). Both our methods have $2*H+2$ more parameters than their baseline (e.g., \ac{LMGC} (\ac{BERT}) has $\approx$ 110M parameters) but equal or fewer parameters compared to related methods (e.g., GlossBERT, KnowBERT W+W). For each polysemous word, we retrieved all possible gloss definitions from WordNet to create sentence-gloss inputs. We increased the hidden dropout probability to $p = 0.2$ as we observed overfitting for most models.
Following\newcite{DevlinCLT19}, we used a batch size of 32 sequences, the AdamW optimizer ($\alpha = 2 \times 10^{-5}, \epsilon = 10^{-5}$), trained three epochs, and chose the best model according to validation loss. We applied the same hyperparameter configuration for all models used in both the SemCor and GLUE benchmarks.

To reduce training time to $\approx \frac{1}{3}$ compared to the approach of \cite{HuangSQH19}, we reduce the sequence length of all models from 512 to 160\footnote{99.8\% of the dataset's examples contain less than 160 tokens; we truncate the remaining sequences to this limit.} as the computational cost of transformers grows quadratic with the sequence length. We treat the class imbalance of positive and negative examples (Table\ref{table:dataset_sem}) with focal loss \cite{LinGGH17} ($\gamma=2$, $\alpha=0.25$).
Our experiments use one NVIDIA Tesla V100 \ac{GPU} with 16GB memory for training and two \ac{GPU}s for evaluation (NVIDIA Tesla V100 and NVIDIA TESLA T4) using the official scorer from\newcite{RaganatoCN17}\footnote{\url{http://lcl.uniroma1.it/wsdeval/}} and Hugginface\footnote{\url{https://tinyurl.com/y4rdkjop}} considering the best results from each evaluation.

We create sentence-gloss inputs by retrieving WordNet's gloss definitions of polysemous words corresponding to the annotated synsets. For \ac{LMGC} and \ac{LMGC-M} (\Cref{sec:lmgc}), we used $8$ candidate glosses in the training procedure as a trade-off between memory requirements and model accuracy, as the mean number of synsets for polysemous words in SemCor is approximately $8.94$ (Table \ref{table:dataset_sem}). If the number of glosses was smaller than $8$, we zero-padded the remaining sequences and disregarded them in the final softmax activation. If the number of glosses exceeded $8$ and the word in question had one synset annotation, we randomly chose $9$ glosses from the set of possible glosses, including the correct one. If the annotation marked multiple glosses as correct, we sampled multiple times from the set of possible glosses to include each sense once in our training procedure. At inference time, we extended $8$ to the number of glosses of each example.
\ac{LMGC} and \ac{LMGC-M} calculate the argmax along with the senses of the softmax activation from \Cref{eq:binary_probs} to predict the correct sense.

To validate that our model is capable of performing \ac{WSD}, while still mastering other tasks, we fine-tuned them on \ac{GLUE}. Therefore, we removed the weight matrices and bias introduced by the heads of each method (i.e., $W_{LMGC}$ and $W_{LMGC-M}$) and represented the input sequence with the final hidden vector $C \in \mathbb{R}^H$ corresponding to the \texttt{[CLS]}-token as the aggregate representation.
For all tasks, except for STS-B, we transformed the embedding into a classification vector applying a new weight matrix $W \in \mathbb{R}^{K \times H}$; where $K$ is the number of labels. For STS-B, we applied a new weight matrix $V \in \mathbb{R}^{1 \times H}$ transforming the aggregate representation into a single regression value. We trained on each \ac{GLUE} task for three epochs with 100 warm-up steps and the same remaining parameters, e.g., learning rate, optimizer, batch size as in SemCor.
\section{Machine-Paraphrased Plagiarism Detection} \label{sec:chap3_method_mpp}
\begin{figure}[!hbt] 
    \centering
    \includegraphics[width=\textwidth]{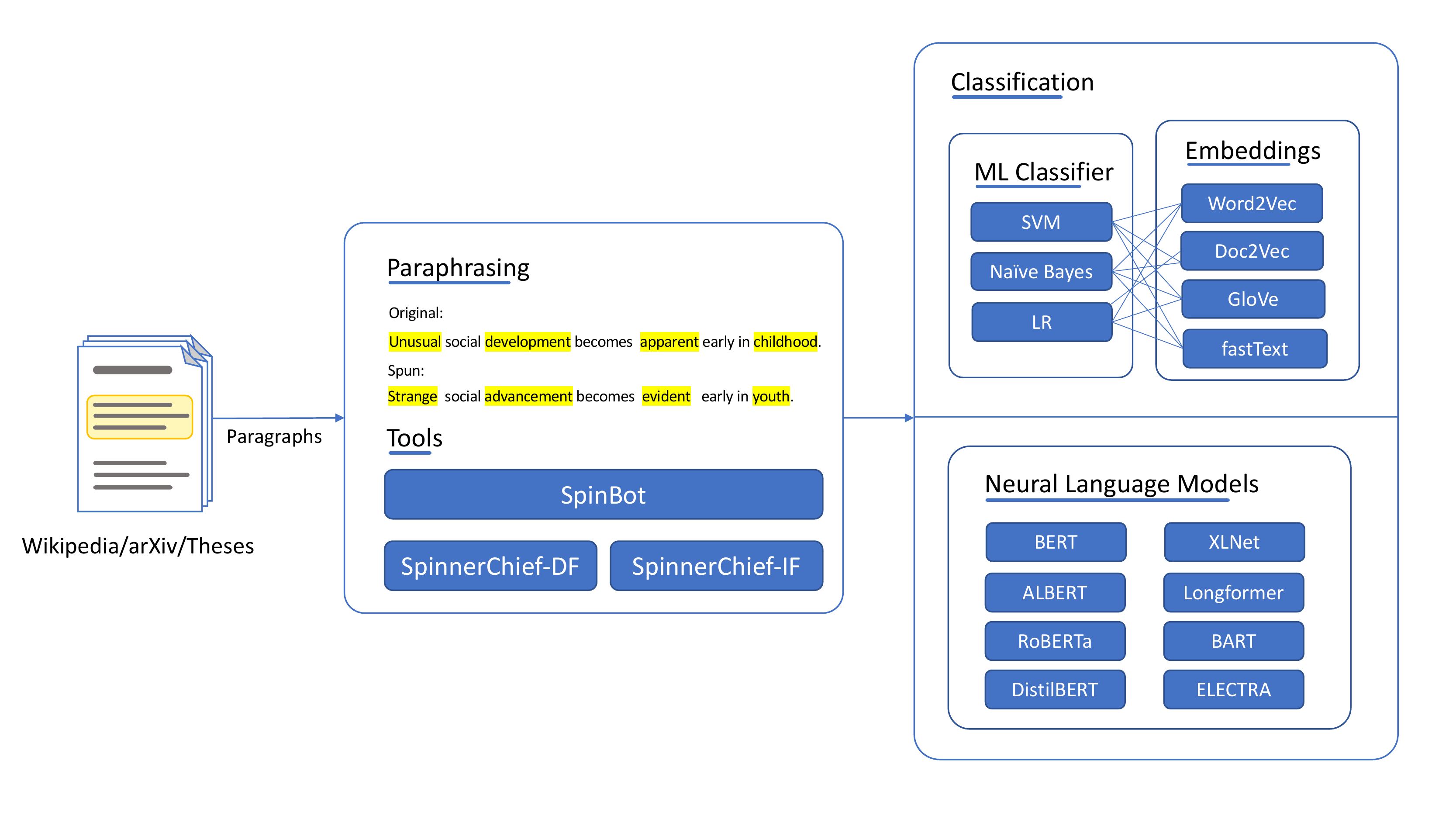}
    \caption[Overview of the Machine-Paraphrased Plagiarism detection methods.]{Overview of the \ac{MPP} detection methods.}
    \label{fig:mpp}
\end{figure}

The \ac{MPP} detection methods presented in the following extend our fellow researchers previous study \cite{FoltynekRSM20} with two new data sources (arXiv and Theses), an additional machine-paraphrase tool (SpinnerChief\footnote{\label{note:spinnerchief}\url{http://www.spinnerchief.com/}}) and eight state-of-the-art neural language models based on the transformers against our best-performing \ac{KD} method. We show the general setup of our study in \Cref{fig:mpp}.

We first perform preliminary experiments with machine learning approaches (e.g., \ac{SVM}, \ac{LR}) to identify the strongest baselines among the paraphrase tools and data sources. Next, we carry out the best performing techniques to be compared to the ones relying on the transformer architecture, which we believe represent the last advancements in \ac{NLP}. We describe the paraphrasing tools, datasets, word embedding models, machine learning classifiers, and neural language models used in our experiments in the following sections.

This comprehensive study aims to evaluate neural language models against the best-performing \ac{KD} method (research \ref{itm:researchtask3}) and to provide a free service that distinguishes human-written from the machine-paraphrased text while being insensitive to the topic, the type of documents, and the paraphrasing tool used. We analyze paragraphs instead of entire documents as it represents a more realistic detection task \cite{RogersonM17, WeberWulff19}. 

\subsection{Paraphrasing Tools} \label{sec:opt}
We employ two commercial paraphrasing services (SpinBot\footnote{\url{https://spinbot.com}} and SpinnerChief\textsuperscript{\ref{note:spinnerchief}}) to obfuscate samples in our training and test sets. We use SpinBot to generate the training and test sets, and SpinnerChief only generates test sets for two reasons. First, SpinnerChief is a paid service making paraphrasing on large datasets expensive. Second, we want to provide a realistic scenario where unseen test cases probably employed a different paraphrasing tool. 

SpinnerChief allows specifying the ratio of words it tries to change. We experimented with two configurations: the default frequency - attempting to change every fourth word (SpinnerChief-DF) and increasing frequency - attempting to change every second word (SpinnerChief-IF). We found the average ratio of replaced words is 20.38\% for SpinBot, 12.58\% for SpinnerChief-DF, and 19.37\% for SpinnerChief-IF when analyzing the paraphrased samples. Thus, the actual frequency with which SpinnerChief replaces words is much lower than its settings suggest. The SpinBot API offers no options to influence the paraphrased text.

\subsection{Training and Testing Datasets} \label{sec:datasets}
Most paraphrasing tools are paid services, which prevents large-scale experimentation. The financial costs and effort required for obtaining and incorporating tool-specific training data would be immense. Therefore, we employed transfer learning, i.e., used pre-trained word embedding models, trained the classifiers in our study on samples paraphrased using SpinBot, and tested whether the classification approach can also identify SpinnerChief's paraphrased text. 

We reused the paragraph training set of our initial study \cite{FoltynekRSM20} as the training set. We paraphrased all 4,012 \textit{featured articles} from the English Wikipedia using SpinBot. We chose featured Wikipedia articles because they objectively cover a wide range of topics in great breadth and depth\footnote{\url{https://en.wikipedia.org/wiki/Wikipedia:Content\_assessment}\label{fn_art-quality}}. Approximately 0.1\% of all Wikipedia articles carry the label featured article. To receive this label, senior Wikipedia editors must confirm the superior quality of the article. Featured articles typically have many authors and revisions. Thus, they are written in high-quality English and unlikely to be biased towards the authors' writing style. 

The training set comprises of $\approx$200.8k paragraphs ($\approx$98.3k original, $\approx$102.5k paraphrased) extracted from $\approx$8k Wikipedia articles. We split each Wikipedia article into paragraphs and discarded those with fewer than three sentences, as our earlier findings showed that such paragraphs often represent titles or irrelevant information \cite{FoltynekRSM20}. 

Our study uses three test sets that we created from preprints of research papers on arXiv, graduation Theses, and Wikipedia articles. \Cref{tab:mpp_test_sets} summarizes the test sets. For generating the arXiv test set, we randomly selected 944 documents from the \textit{no problems} category of the arXiv project\footnote{\url{https://kwarc.info/projects/arXMLiv/}}. The 
Wikipedia test set is identical to the one in our preliminary study \cite{FoltynekRSM20}. The paragraphs in the test set were generated analogously to the training set. The Theses test set comprises the paragraphs in 50 randomly selected graduation Theses of \textit{English as a second language} students at the Mendel University in Brno, Czech Republic. The Theses are from a wide range of disciplines, e.g., economics, agronomy, forestry, and computer science, and cover all academic levels (i.e., B.Sc., M.Sc., and Ph.D.). Unlike the arXiv and Wikipedia documents, the Theses were only available as PDF files, thus having to be converted to plain text. We removed all content before the introduction section of each thesis, the bibliography, and all appendices because these parts of the PDF tended to produce content without valuable semantic information.

\begin{table}[htb]
\caption[Overview of the Machine-Paraphrased Plagiarism test sets.]{Overview of the \ac{MPP} test sets.}
\centering
\resizebox{\textwidth}{!}{
    \begin{tabular}{lrrrrrr}
    \toprule
    \textbf{No. of paragraphs} & \multicolumn{2}{c}{\textbf{arXiv}}                                                                         & \multicolumn{2}{c}{\textbf{Theses}}                                                                        & \multicolumn{2}{c}{\textbf{Wikipedia}}                                                                     \\
                               & \multicolumn{1}{c}{Original} & \multicolumn{1}{c}{\begin{tabular}[c]{@{}c@{}}Paraphrased\end{tabular}} & \multicolumn{1}{c}{Original} & \multicolumn{1}{c}{\begin{tabular}[c]{@{}c@{}}Paraphrased\end{tabular}} & \multicolumn{1}{c}{Original} & \multicolumn{1}{c}{\begin{tabular}[c]{@{}c@{}}Paraphrased\end{tabular}} \\ \midrule
    \textbf{SpinBot}           & 20.97k                       & 20.87k                                                                      & 5.23k                        & 3.46k                                                                       & 39.26k                       & 40.73k                                                                      \\
    
    \textbf{SpinnerChief-DF}   & 20.97k                       & 21.72k                                                                      & 2.38k                        & 2.94k                                                                       & 39.26k                       & 39.7k                                                                      \\ 
    \textbf{SpinnerChief-IF}   & 20.97k                       & 21.67k                                                                      & 2.38k                        & 2.94k                                                                       & 39.26k                       & 39.62k                                                                      \\ \bottomrule
    \end{tabular}%
}
\label{tab:mpp_test_sets}
\end{table}

\subsection{Word Embedding Models} \label{sec:method_embeddings}
\Cref{tab:mpp_word_embed_det} summarizes the word embedding models used in our experiments: \ac{GloVe}\footnote{\url{https://nlp.stanford.edu/projects/glove/}} \cite{PenningtonSM14}, \ac{W2V}\footnote{\url{https://code.google.com/archive/p/word2vec/}} \cite{MikolovSCC13}, \ac{FT}\footnote{\url{https://fasttext.cc/docs/en/english-vectors.html}}(\ac{FT}-rw and \ac{FT}-sw) \cite{BojanowskiGJM17}, and \ac{D2V} \cite{LeM14} which was trained from scratch. Following the hyperparameter recommendations of \newcite{LauB16} for general-purpose applications, \ac{D2V} uses uses a distributed bag-of-words training, a window size of 15 words, a minimum count of five words, trained word-vectors in skip-gram fashion, averaged word vectors, and 30 epochs. All word embedding models have 300 dimensions. Parameters we do not explicitly mention correspond to the default values in the gensim API\footnote{\url{https://radimrehurek.com/gensim/models/doc2vec.html}}.

Our rationale for choosing the pre-trained word embedding models was to explore the most prominent techniques regarding their suitability for the plagiarism detection task. \ac{GloVe} \cite{PenningtonSM14} builds a co-occurrence matrix of the words in a corpus and explores the ratio of the word probabilities in a text to derive its semantic vectors as a count-based model. \ac{W2V} \cite{MikolovSCC13} training tries to predict either a word given its context (cbow) or the context given the word (skip-gram). 

Even though \ac{GloVe} and \ac{W2V} are routinely applied for numerous \ac{NLP} tasks \cite{ConneauKSB17, RuasGA19, RuasFGd20}, they do not consider two important linguistic characteristics: word ordering and sub-wording. To explore these characteristics, we also included \ac{FT} \cite{BojanowskiGJM17} and \ac{D2V} \cite{LeM14}. \ac{FT} builds its word representation by extending the skip-gram model with the sum of the n-grams of its constituent sub-word vectors. For the \ac{D2V} model, two training options exist -- named distributed memory (pv-dm) and distributed bag-of-words (pv-dbow). The former is akin to \ac{W2V}'s cbow, while the latter is related to skip-gram. Both options introduce a new paragraph-id vector that is updated for each context window on every timestamp. The paragraph-id vector seeks to capture the semantics of the embedded object. We chose a pv-dbow over pv-dm model because of its results in semantic similarity tasks \cite{LauB16}.

\begin{table}[htb]
\caption[Word embedding models for the Machine-Paraphrased Plagiarism detection experiments.]{Word embedding models for the \ac{MPP} detection experiments.}     
\centering
\resizebox{\textwidth}{!}{
\begin{tabular}{lll}\toprule
\textbf{Algorithm} & \textbf{Main Characteristics}                      & \textbf{Training Corpus}                     \\ \midrule
GloVe     & Word-word co-occurrence matrix            & Wikipedia Dump 2014 + Gigaword 5    \\
W2V  & Continuous Bag-of-Words                   & Google News                        \\
pv-dbow   & Distributed Bag-of-Words    &Wikipedia Dump 2010 \\
fastText-rw  & Skip-gram without sub-words                                  & Wikipedia Dump 2017 + UMBC     \\
fastText-sw  & Skip-gram with sub-words                                 & Wikipedia Dump 2017 + UMBC         \\ \bottomrule
\end{tabular}
}
\label{tab:mpp_word_embed_det}
\end{table}

In our experiments, we represent each text as the average of its constituent word vectors by applying the word embedding models in \Cref{tab:mpp_word_embed_det}. All models, except for \ac{D2V}, yield a vector representation for each word. In \ac{D2V}, the embedded tokens represent the entire text. Thus, a match between a text not part of the external training corpus and the pre-trained \ac{D2V} model is unlikely. Inferring the vector representations for unseen texts requires an additional training step with specific parameter tuning. For all texts in our test sets, we performed this extra step using the following hyperparameters for the gensim API: $\alpha=10^{-4}, \alpha_{min} = 10^{-6}$, and $300$ epochs \cite{LauB16}. The resulting pv-dbow embedding model requires at least 7 GB of memory, compared to 1-3 GB required for other word embedding models. The higher memory consumption of pv-dbow can make it unsuitable for some use cases.

\subsection{Machine Learning Classifiers} \label{sec:classifiers}
After applying the pre-trained models to our training and test sets, we forward the embeddings to three machine learning classifiers: \ac{LR}, \ac{SVM}, and \ac{NB}. We use multiple classifiers to explore the stability of the word embedding models concerning each classifier's characteristics. We employ a grid-search approach for finding the optimal parameter values for each classifier as\Cref{tab:mpp_gridsearch} shows.

\begin{table}[htb!] \label{tab:mpp_gridsearch}
\centering
\caption[Grid-search parameter for Machine Learning classifiers.]{Grid-search parameter for \ac{ML} classifiers considering the scikit-learn\footnotemark package in Python.}
\resizebox{.9\textwidth}{!}{
  \begin{tabular}{llr}
  \toprule
    Classifier                                                                        & Parameter             & Range                                       \\\midrule
    \multirow{4}{*}{\begin{minipage}{0.7in}Logistic\\ Regression\end{minipage}}       & solver                & newton-cg, lbfgs, sag, saga                 \\
                                                                                      & maximum iteration     & 500, 1000, 1500                             \\
                                                                                      & multi-class           & ovr, multinomial                            \\
                                                                                      & tolerance             & 0.01, 0.001, 0.0001, 0.00001                \\\cmidrule(lr){2-3}
    \multirow{4}{*}{\begin{minipage}{0.5in}Support\\ Vector\\Machine\end{minipage}}   & kernel                & linear, radial bases function, polynomial   \\
                                                                                      & gamma                 &  0.01, 0.001, 0.0001, 0.0001                \\
                                                                                      & polynomial degree     & 1, 2, 3, 4, 5, 6, 7, 8, 9                   \\
                                                                                      & C                     & 1, 10, 100                                  \\
    \bottomrule
  \end{tabular}
}
\end{table}
\footnotetext{\url{https://scikit-learn.org}}

\subsection{Neural Language Models} \label{sec:neuralapproaches}
We use the following neural language models based on the Transfomer architecture in our experiments: \ac{BERT} \cite{DevlinCLT19}, \ac{RoBERTa} \cite{LiuHCG19}, \ac{ALBERT} \cite{LanCGG19}, DistilBERT \cite{SanhDCW19}, ELECTRA \cite{ClarkLLM20}, BART \cite{LewisLGG19}, XLNet \cite{YangDYC19}, and Longformer \cite{BeltagyPC20}.
Our rationale for choosing these models was two-fold. First, we explore models closely related or based on \ac{BERT}, either by improving it through additional training time and data (\ac{RoBERTa}) or compressing the architecture with minimal performance loss (DistilBERT, \ac{ALBERT}). Second, we used contrasting models to \ac{BERT}, that although relying on the transformer architecture, significantly change the training objective (XLNet), the underlying attention mechanism (Longformer), or employ a discriminative learning approach (ELECTRA, BART).

To distinguish between a paraphrased and human-written paragraph, a randomly initialized linear layer on top of the model's embedding of the \texttt{[CLS]}-token performs binary classification with cross-entropy loss.
For all models, we use the base version and the official pre-trained weights with the following configurations: a sequence length of $512$ tokens, an accumulated batch size of $32$, the Adam optimizer with a learning rate of $\alpha = 2\times 10^{-5}$, $\beta_1 = 0.9$, $\beta_2 = 0.999$, $\epsilon = 10^{-8}$, and PyTorch's native automated mixed-precision format. Using a common sequence length of 512 tokens allows for a fair comparison between models without losing important context information\footnote{99.35\% of the dataset's text can be described entirely with less than 512 tokens.}. 
In \Cref{sec:eval_nlm} we provide more details about these models characteristics.
\chapter{Evaluation} \label{Chapter4}
This chapter evaluates our presented \ac{KD} methods in the tasks of language modeling, \ac{WSD}, \ac{NLU}, and \ac{PD} with a total of 20 tasks.
First, we provide evidence that our weighted multi-teacher \ac{KD} reduces pre-training time and obtains superior validation performance over large-scale teachers when the dataset size and computational budget remains equal between the models (\Cref{sec:eval_lm}).

Next, we evaluate methods to incorporate lexical knowledge into \ac{LM}s on five \ac{WSD} benchmarks, and eight \ac{NLU} tasks (\Cref{sec:chap4_eval_wsd}). 
We show \ac{LMGC} obtains superior performance in the \ac{WSD} task over related work when using XLNet as the backbone network while keeping the same number of parameters. When transferred to \ac{NLU} tasks, \ac{LMGC-M} performs superior on average over \ac{BERT}, validating its improved semantic representations. The code\footnote{\url{https://github.com/jpwahle/word-sense-disambiguation}}, and pre-trained models\footnote{\label{footnote:chap4_huggingface}\url{https://huggingface.co/models?search=jpwahle}} to reconstruct our experiments are publicly available.

Finally, we apply our lexical \ac{KD} models to the task of \ac{MPP} detection. We compare our method to eight neural \ac{LM}s and three \ac{ML} techniques using word-embeddings on encyclopedia articles, student Theses, and research papers (\Cref{sec:chap4_eval_mpp}). We show \ac{ML} methods are superior over a human baseline and accurately predict plagiarism when the data source and paraphrasing tool changes. Compared to two popular \ac{PD} systems, neural language models detect plagiarism on new sources more accurately, which makes them a suitable extension to \ac{PD} systems. We release the data\footnote{\url{https://doi.org/10.5281/zenodo.3608000}}, code\footnote{\url{https://https://github.com/jpwahle/iconf22-paraphrase}}, and pre-trained models\textsuperscript{\ref{footnote:chap4_huggingface}} of our study, as well as a web-based demonstration system\footnote{\url{http://purl.org/spindetector}}.

\section{Language Modeling} \label{sec:eval_lm}
When evaluating a \ac{LM}s capability of modeling the probability of words, \ac{PPL} \cite{JelinekMBB77} is a widely used measure \cite[Chapter 3]{Jurafsky09}. \ac{PPL} is the inverse probability normalized by the number of words in the test corpus as \Cref{eq:lm_perplexity} shows.

\begin{align} \label{eq:lm_perplexity}
    PPL(w_1,...,w_n) &= P(w_1,...,w_n)^{\frac{1}{n}} \\
    &=\sqrt[n]{\frac{1}{\prod_{i=1}^n P(w_i \mid w_{<i})}}
\end{align}

To calculate \ac{PPL}, the \ac{LM} needs to assign conditional probabilities to consecutive tokens. As masked \ac{LM}s predict the probability of tokens for corrupted sequences rather than succeeding tokens, the \ac{PPL} is not well defined.
However, when using the cross-entropy loss for \ac{MLM}, \Cref{eq:perplexity_cross_entropy} shows that perplexity is proportional to the cross-entropy loss by an exponential function.

\begin{align} \label{eq:perplexity_cross_entropy}
    CE(w1, ..., w_n) &= -\frac{1}{n} \log(P(w_1,...,w_n)) \\
    PPL(w_1, ..., w_n) &= e^{CE(w_1,...,w_n)}
\end{align}

We use cross-entropy to measure a relative gain in language modeling performance during training between the student and teacher models.
Therefore, we use the validation loss --- which is the cross-entropy for token predictions --- to obtain a measure proportional to the inverse probability. This measure provides us with evidence about the language modeling capabilities of the model over the validation corpus. 

Another robust way to compare \ac{LM}s is the validation on general \ac{NLU} tasks.
We evaluate the methods proposed in \Cref{sec:chap3_method_wsd} in the task of \ac{WSD} and on a variety of \ac{NLU} tasks.
Due to the computational requirements of large-scale models, we leave their investigation for future work.

\subsection{Results and Discussion}

\begin{figure}
    \centering
    \includegraphics[width=.8\textwidth]{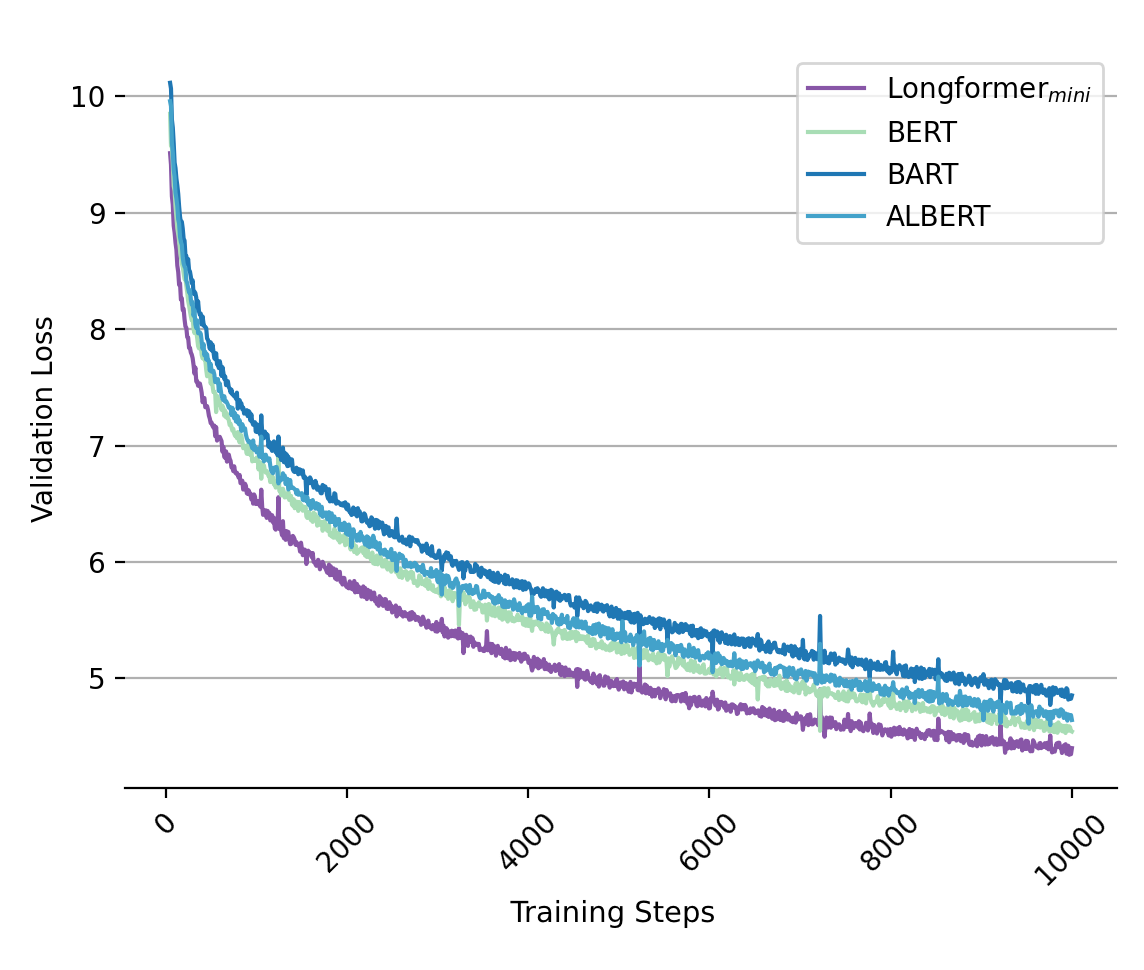}
    \caption[Validation loss by the number of training steps for teacher and student models.]{Validation loss by the number of training steps for student (Longformer$_{mini}$) and teacher (\ac{BERT}, \ac{ALBERT}, BART) models. The loss is a smoothed version of the accumulated cross-entropy of \ac{MLM}.}
    \label{fig:kd_val_loss}
\end{figure}

\Cref{fig:kd_val_loss} reports the validation cross-entropy loss on the ground truth targets for the student model (Longformer$_{mini}$) and each teacher model (\ac{BERT}, \ac{ALBERT}, BART) over the number of training steps. Each training step equals $256 \times 512$ tokens. Each teacher model curve shows the validation loss for training with the official training procedure and randomly initialized weights.

Our first observation is that the student model Longformer$_{mini}$ initially converges faster than its larger counterparts. We hypothesize the initialization of layers with pre-trained ones increases the convergence rate compared to randomly initialized weights. Furthermore, the weighted target distribution includes highly confident predictions of the pre-trained teachers. As we use similar datasets to the ones used to train the teacher models, we assume confident predictions are also accurate with high probability. 

The validation loss of all teachers behaves similarly, with \ac{BERT} performing marginally better than \ac{ALBERT}. BART's validation loss converges slower than the other models which we suppose results from BART being a combination of \ac{AE} and \ac{AR} models resulting in more training time to converge to both objectives. We assume different levels of noise in the validation loss at different training steps per model appear due to randomly shuffling the training data leading to batches that sometimes do not represent the data distribution well.

At later training steps (starting at step $\approx$6000), the student model (Longformer$_{mini}$) is still slightly superior over its teachers, but its advantage in validation loss decreased, supporting our hypothesis about improvements by weight initializations. However, we can interpret the decrease in learning steepness in two ways. First, the student might already have captured a large proportion of the knowledge encoded in the teachers and reached its capacity. Considering the continuation of training, we expect to observe overfitting by a slow change in the curve's direction.
Second, the advantage of Longformer$_{mini}$'s layer initialization does not impact the training convergence much in advanced training steps (step $\approx$6000) and its learning steepness converges to the teacher models.

\subsection{Limitations}
Although the Longformer$_{mini}$ reduced its validation loss faster than its teacher models, we must consider more experiments to conclude its converged model performs better in language modeling than its teachers. These extending experiments must include converging Longformer$_{mini}$ regarding its validation loss and testing the resulting model on \ac{NLU} benchmarks against its teachers and other \ac{KD} methods.

We aim to converge the model in future experiments and test its generalization on \ac{NLU} benchmarks to provide an efficient replacement for large models to the community.
In the following, we focus on more promising \ac{KD} methods using lexical sources that are cheaper to train and prove high performance in \ac{WSD}, \ac{NLU}, and \ac{PD} as the total computational budgets for this thesis are limited.
\section{Word Sense Disambiguation and Natural Language Understanding} \label{sec:chap4_eval_wsd}
We evaluate our proposed lexical \ac{KD} methods on two large scale benchmarks, namely SemCor 3.0 \cite{MillerLTB93} and the General Language Understanding Evaluation (GLUE) \cite{WangSMH19}. Comparing \ac{LMGC} and \ac{LMGC-M} with state-of-the-art systems, we provide an analysis and discussions of experimental results and point out possible limitations of the presented methods.

\subsection{Results \& Discussion}
\Cref{table:model_comparison} reports the results of applying \ac{LMGC} to different transformer models. Our rationale for choosing the models was two-fold. First, we explore models closely related to or based on \ac{BERT}. Either the chosen models improve \ac{BERT} through additional training time and data (\ac{RoBERTa}), or compress the architecture with minimal performance loss using parameter optimizations or \ac{KD} (\ac{ALBERT}, DistilBERT). Second, we chose models that significantly change the training objective (XLNet) or employ a discriminative learning approach (ELECTRA, BART).
In \Cref{tab:semeval}, we compare our techniques to other contributions in \ac{WSD}. We report all results of SemCor according to \newcite{RaganatoCN17}.

\begin{table}[htb]
\caption[SemCor test results of LMGC for base transformer models]{SemCor test results (F1) of \ac{LMGC} for base transformer models. \textbf{Bold} font indicates the best results.}
\centering
\resizebox{.75\textwidth}{!}{ 
    \begin{tabular}{lcccccc} 
    \toprule
    \textbf{System} & \textbf{SE7} & \textbf{SE2} & \textbf{SE3} & \textbf{SE13} & \textbf{SE15} & \textbf{All} \\ \midrule
    \ac{BERT} (\citeyear{DevlinCLT19}) & 71.9 & 77.8 & 74.6 & 76.5 & 79.7 & 76.6 \\\hdashline
    \ac{RoBERTa} (\citeyear{LiuOGD19}) & 69.2 & 77.5 & 73.8 & 77.2 & 79.7 & 76.3 \\
    DistilBERT (\citeyear{SanhDCW19}) & 66.2 & 74.9 & 70.7 & 74.6 & 77.1 & 73.5  \\
    \ac{ALBERT} (\citeyear{LanCGG19}) & 71.4 & 75.9 & 73.9 & 76.8 & 78.7 & 75.7 \\
    BART (\citeyear{LewisLGG19}) & 67.2 & 77.6 & 73.1 & 77.5 & 79.7 & 76.1 \\
    XLNet (\citeyear{YangDYC19}) & \textbf{72.5} & \textbf{78.5} & \textbf{75.6} & \textbf{79.1} & \textbf{80.1} & \textbf{77.2} \\
    ELECTRA (\citeyear{ClarkLLM20}) & 62.0 & 71.5 & 67.0 & 73.9 & 76.0 & 70.9 \\
    \bottomrule
    \end{tabular}
   }
\label{table:model_comparison}
\end{table}

\ac{RoBERTa} shows inferior F1 than the baseline model \ac{BERT} although it uses more data and training time. DistilBERT and \ac{ALBERT} perform worse than \ac{BERT}, which we expected given that they use significantly fewer parameters. However, \ac{ALBERT} achieves good results with only $\approx$ 10\% of \ac{BERT}'s parameters. ELECTRA and BART results show the models' denoising approach is not suitable for our \ac{WSD} setup. Besides, BART achieves similar performance as \ac{BERT} but uses ~26\% more parameters. XLNet continually performs better than \ac{BERT} on all evaluation sets while using almost the same number of parameters. Therefore we selected XLNet for our models' variation.

\begin{table}[htb]
\caption[Model size comparison on the three largest SemCor evaluation sets.]{Classification results (F1) of two model sized of \ac{BERT} ($base$ and $large$) on the three largest SemCor evaluation sets.}
\centering
    \resizebox{.5\textwidth}{!}{ 
    \begin{tabular}{l ccc} \toprule
    \textbf{System} & \textbf{SE2} & \textbf{SE13} & \textbf{All} \\ \midrule
    BERT$_{base}$ \citeyear{DevlinCLT19} & 77.78 & 76.52 & 76.59 \\
    BERT$_{large}$ \citeyear{DevlinCLT19} & 78.26 & 76.46 & 76.67 \\
    \bottomrule
    \end{tabular}
    }
\label{table:base_large}
\end{table}

We did not consider larger models than the base configuration as our experiments showed a difference of 0.08\% in F1 between \ac{BERT}$_{base}$ and \ac{BERT}$_{large}$ (see \Cref{table:base_large}) for the SemCor datasets, which is in line with \newcite{BlevinsZ20}'s findings. Thus, we consider the base configuration as sufficient for our experiments.

\begin{table}[!b]
\caption[Classification results on the SemCor test sets.]{Classification results (F1) on the SemCor test sets compared to state-of-the-art techniques. \textbf{Bold} font indicates the best results.}
\centering
 \resizebox{.8\textwidth}{!}{ 
    \begin{tabular}{lcccccc} \toprule
    \textbf{System} & \textbf{SE7} & \textbf{SE2} & \textbf{SE3} & \textbf{SE13} & \textbf{SE15} & \textbf{All} \\ \midrule
    CAN (\citeyear{LuoLHX18})                   &  - & 72.2 & 70.2 & 69.1 & 72.2 & 70.9 \\
    HCAN (\citeyear{LuoLHX18})                    & - & 72.8 & 70.3 & 68.5 & 72.8 & 71.1\\
    LMMS$_{BERT}$ (\citeyear{LoureiroJ19})                              & 68.1 & 76.3 & 75.6 & 75.1 & 77.0 & 75.4 \\
    GLU (\citeyear{HadiwinotoNG19})        & 68.1 & 75.5 & 73.6 & 71.1 & 76.2 & 74.1 \\    
    GlossBERT (\citeyear{HuangSQH19})        & 72.5 & 77.7 & 75.2 & 76.1 & 80.4 & 77.0 \\ 
    BERT$_{WSD}$  (\citeyear{DuQS19})                          & - & 76.4 & 74.9 & 76.3 & 78.3 & 76.3 \\
    KBERT-W+W (\citeyear{PetersNLS19})           & - & - & - & - & - & 75.1 \\ \hdashline
    \ac{LMGC} (\ac{BERT}) & 71.9 & 77.8 & 74.6 & 76.5 & 79.7 & 76.6 \\
    \ac{LMGC-M} (\ac{BERT}) & 72.9 & 78.2 & 75.5 & 76.3 & 79.5 & 77.0 \\
    \ac{LMGC} XLNet & 72.5 & 78.5 & 75.6 & 79.1 & 80.1 & 77.2 \\
    \ac{LMGC-M} XLNet & \textbf{73.0} & \textbf{79.1} & \textbf{75.9} & \textbf{79.0} & \textbf{80.3} & \textbf{77.5} \\
    \bottomrule
    \end{tabular}
}
\label{tab:semeval}
\end{table}

\Cref{tab:semeval} shows an overall improvement when comparing \ac{LMGC} to related approaches. \ac{LMGC} (BERT) generally outperforms \newcite{DuQS19}'s baseline approach (BERT$_{WSD}$) and KBERT-W+W, which has four times the number of parameters. We outperform GlossBERT in all test sets by using an optimal transformer (XLNet) and adjustments in the training procedure \ac{LMGC-M}. We exclude \ac{EWISER} \cite{BevilacquaN20} which explores additional knowledge other than gloss definitions (e.g., knowledge graphs). We leave for future work the investigation of \ac{BEM} \cite{BlevinsZ20}, a recently published bi-encoder with separate encoders for context and gloss.

\begin{table}[htb]
\caption[Classification results for LMGC and LMGC-M, scored by the GLUE evaluation benchmark.]{Test results, scored by the \ac{GLUE} evaluation benchmark. As in \ac{BERT}, we exclude the problematic WNLI set. We report F1-score for QQP and MRPC, \ac{SC} for STS-B, \ac{MC} for CoLA, and \ac{ACC} for the other tasks (with matched/mismatched accuracy for MNLI). \textbf{Bold} face indicates the best result per dataset.}
\centering
 \resizebox{\textwidth}{!}{
    \begin{tabular}{lccccccccc} \toprule
    \multirow{3}{*}{\textbf{System}} & \multicolumn{2}{c}{\textbf{Classification}} & \multicolumn{3}{c}{\textbf{Semantic Similarity}} & \multicolumn{3}{c}{\textbf{Natural Language Inference}} & \multicolumn{1}{c}{\textbf{Average}} \\
    \cmidrule(lr){2-3}
    \cmidrule(lr){4-6}
    \cmidrule(lr){7-9}
    \cmidrule(lr){10-10}
    {} & CoLA & SST-2 & MRPC & STS-B & QQP & MNLI & QNLI & RTE & - \\
    {} & (mc) & (acc) & (F1) & (sc) & (F1) & m/mm(acc) & (acc) & (acc) & - \\ \midrule
    BERT$_{base}$ & 52.1 & 93.5 & \textbf{88.9} & 85.8 & 71.2 & 84.6/83.4 & \textbf{90.5} & 66.4 & 79.6  \\
    GlossBERT & 32.8 & 90.4 & 75.2 & 90.4 & 68.5 & 81.3/80 & 83.6 & 47.3 & 70.7 \\ \hdashline
    LMGC (\ac{BERT}) & 31.1 & 89.2 & 81.9 & 89.2 & 87.4 & 81.4/80.3 & 85.4 & 60.2 & 74.5 \\
    LMGC-M (\ac{BERT}) & \textbf{55.0} & \textbf{94.2} & 87.1 & \textbf{88.1} & \textbf{90.8} & \textbf{85.3/84.2} & 90.1 & \textbf{69.7} & \textbf{82.5}  \\
    \bottomrule
    \end{tabular}
    }
\label{table:glue}
\end{table}

We evaluate our suspicion that \ac{WSD} training allows language models to achieve higher generalization. Hence, we fine-tune the weights from our converged models in general language tasks from GLUE \cite{WangSMH19}. 
\Cref{table:glue} shows the results of our proposed methods against the previously best-performing model in \ac{WSD} (GlossBERT \cite{HuangSQH19}), and the official \ac{BERT}$_{base}$ model on the \ac{GLUE} datasets. \ac{LMGC-M} achieves a 2.9\% increase in performance on average over \ac{BERT}. Although \ac{LMGC} and GlossBERT perform well in \ac{WSD}, they cannot maintain good performance on other \ac{GLUE} tasks. Still, \ac{LMGC} performs better than GlossBERT, which we assume is due to its improved loss function regarding negative examples. \ac{LMGC-M} outperforms \ac{BERT} on most tasks and is certainly comparable to the others. Therefore, we conclude the incorporation of \ac{MLM} adds value from \ac{WSD} into natural language understanding. We exclude XLNet from the comparison to show that the additional performance is attributable to our method, not to the improvement of XLNet over BERT.
In this work, we did not compare \ac{LMGC} and \ac{LMGC-M} to the other \ac{WSD} methods performing worse than \newcite{HuangSQH19} in the \ac{WSD} task (\Cref{tab:semeval}) because reconstructing their models is computationally expensive.

\subsection{Limitations} \label{sec:limitations}
\ac{LMGC-M} optimizes two objectives simultaneously, which is more expensive than \ac{LMGC}. However, \ac{LMGC-M} has the same number of parameters as \ac{LMGC} during testing, resulting in approximately the same inference time.

Furthermore, we seek to investigate the impact of each component of our presented methods (e.g., weak supervision signals, parallel predictions, short sequence lengths) by extending the ablation. We leave this study to future work as it requires experimenting with the models with many of their variations.
\section{Machine-Paraphrased Plagiarism Detection} \label{sec:chap4_eval_mpp}
We evaluate the effectiveness of the classification approaches in identifying machine-paraphrased text with three experiments. \Cref{sec:eval_auto} presents the results of applying word embedding models with machine learning classifiers and neural language models to text from Wikipedia, arXiv, and Theses modified with two paraphrasing tools (SpinBot, SpinnerChief). \Cref{sec:eval_human,sec:eval_turnitin} establish two baselines for the results of the automated classification approaches by indicating how well human experts and two text-matching systems identify machine-paraphrased text, respectively.

\subsection{Automated Classification} \label{sec:eval_auto}
\Cref{tab:class-results-sb,tab:class-results-sc-all,tab:class-results-nlm} show the micro-averaged F1 scores (\mbox{F1-Micro}) for identifying paraphrased paragraphs using SpinBot and SpinnerChief (DF and IF), for machine learning and transformer-based techniques.
We show the best performing combination of embedding and classifiers in the row of the test set name for the machine learning approaches. We use the top-ranked results (\Cref{tab:class-results-sb,tab:class-results-sc-all}) as a baseline against the transformer-based ones in \Cref{tab:class-results-nlm}.

\subsubsection{Machine Learning Results for SpinBot}
\ac{GloVe} in combination with \ac{SVM} achieved the best classification performance for all test sets (\Cref{tab:class-results-sb}). The combination of \ac{W2V} and \ac{SVM} performed nearly as good as \ac{GloVe}+\ac{SVM} for all test sets. For Theses and Wikipedia, the performance difference between \ac{GloVe}+\ac{SVM} and \ac{W2V}+\ac{SVM} was less than 2\%, but for arXiv the difference was considerably higher with 6.66\%. All word embedding models achieve their best results for Wikipedia test cases. This result corresponds to our expectation, since all pre-trained word embedding models, except for \ac{W2V}, use Wikipedia dumps as part of their training corpora.

\begin{table}[htb!]
\caption[Classification results of Machine Learning methods for SpinBot. ]{Classification results (F1-Micro) for SpinBot. \textbf{Bold} face indicates the best score for each combination of dataset and classifier configuration. The highest scores for each classifier are repeated in the dataset name row.}
\centering
\resizebox{.55\textwidth}{!}{
\begin{threeparttable}
\centering
\begin{tabular}{lrrrrr}\toprule
                 & GloVe    & W2V      &  D2V       & FT-rw     & FT-sw    \\\midrule
\textbf{arXiv}           &\textbf{86.46}	&79.80	&72.40	&78.40	&74.14  \\ \midrule
LR              &76.53	&74.82	&69.42	&75.08	&65.92  \\
SVM	            &86.46	&79.80	&72.40	&76.31	&74.15	\\
NB	            &79.17	&74.23	&57.99	&78.40	&64.96	\\\midrule

\textbf{Theses}          &\textbf{83.51}	&81.94	&61.92	&72.75	&64.78  \\ \midrule
LR              &68.55	&72.89	&59.97	&69.17	&64.03  \\
SVM	            &83.51	&81.94	&61.92	&72.75	&64.78	\\
NB	            &75.22	&74.18	&42.30	&72.11	&61.99	\\\midrule

\textbf{Wikipedia}       &\textbf{89.55}	&87.27	&83.04	&86.15	&82.57  \\ \midrule
LR              &80.89	&84.50	&81.08	&85.13	&78.97  \\
SVM	            &89.55	&87.27	&83.04	&86.15	&82.57	\\
NB	            &69.68	&69.84	&58.88	&70.05	&64.47	\\
\bottomrule
\end{tabular}
\end{threeparttable}
}
\label{tab:class-results-sb}
\end{table}

While all classification approaches performed worse for Theses, the drop in performance was smaller than we expected. The F1-Micro score of the best approach for Theses (\ac{GloVe}+\ac{SVM}) is 6.04\% lower than for Wikipedia and 3.09\% lower than for arXiv. All embedding models, except for \ac{W2V}, perform worse for Theses than for arXiv. This finding suggests writing quality in student Theses mildly affects the detection of machine-paraphrased text.  

Although \ac{D2V} seeks to mitigate the shortcomings of its predecessor \ac{W2V}, such as word order and variable-length encoding, \ac{W2V} surpassed \ac{D2V} for all test sets. One reason for this observation can be the comparably short length of the paragraphs we consider. \newcite{LauB16} found \mbox{\ac{D2V}'s} performance decreases for short documents. The results for paragraphs in \Cref{tab:class-results-sb} and for documents in our preliminary study \cite{FoltynekRSM20}, where \ac{D2V} was the best-performing approach, agree with this finding. 

When considering \ac{FT} in \Cref{tab:class-results-sb}, we observe the same behavior as for \ac{W2V} and \ac{D2V}. In theory, the pre-trained \ac{FT}-sw model is supposed to capture the nuances of the words inner components, i.e., sub-word embeddings. Therefore, we expected a better performance of \ac{FT}-sw compared to \ac{FT}-rw, which encodes whole words. However, \ac{FT}-rw and simpler models, i.e., \ac{GloVe} and \ac{W2V}, performed better than \ac{FT}-sw for all test sets. These results are interesting as both \ac{FT}-rw and \ac{FT}-sw have been trained using the same corpus.

\subsubsection{Machine Learning Results for SpinnerChief} 
For test cases using SpinnerChief, we observed a decrease in the classification performance compared to classifying SpinBot test cases. We expected this decrease as SpinnerChief was not used to train these models. The average decrease in the F1-Micro scores was approximately 17\% when using SpinnerChief's default setting of attempting to replace every fourth word (\Cref{tab:class-results-sc-all}) and approximately 13\% for increasing the frequency of attempted word replacements to every other word (\Cref{tab:class-results-sc-all}).

\begin{table*}[htb!]
\caption[Classification results of Machine Learning methods for SpinnerChief.]{Classification results (F1-Micro) for SpinnerChief. DF - default frequency - attempting to change every fourth word IF - increased frequency - attempting to change every second word. \textbf{Bold} face indicates the best score for each combination of dataset, and classifier configuration. The highest scores for each classifier are repeated in the dataset name row.}
\centering
\resizebox{\textwidth}{!}{
\begin{tabular}{lcccccccccc}\toprule
\multirow{2}{*}{} &\multicolumn{5}{c}{\textbf{SpinnerChief-DF}} & \multicolumn{5}{c}{\textbf{SpinnerChief-IF}}  \\
\cmidrule(lr){2-6}
\cmidrule(lr){7-11}
                & GloVe    & W2V      &  D2V       & FT-rw     & FT-sw    & GloVe    & W2V      &  D2V       & FT-rw     & FT-sw\\
\cmidrule(lr){1-6}
\cmidrule(lr){7-11}

\textbf{arXiv}           &58.48	&\textbf{59.78}	&56.46	&57.42	&59.72   &64.34	&\textbf{65.89}	&59.27	&63.70	&63.66  \\ 

\cmidrule(lr){1-6}
\cmidrule(lr){7-11}
LR              &52.14	&55.43	&56.46	&57.42	&58.64  &54.92	&59.61	&59.07	&61.74	&61.57  \\
SVM	            &58.42	&57.65	&56.43	&56.43	&59.72	 &64.12	&62.77	&59.27	&62.97	&63.66  \\
NB	            &58.48	&59.78	&51.58	&51.58	&55.21	&64.34	&65.89	&52.21	&63.70	&59.33  \\ 
\cmidrule(lr){1-6}
\cmidrule(lr){7-11}

\textbf{Theses}          &52.63	&53.60	&\textbf{59.09}	&53.08	&57.25   &58.57	&58.24	&\textbf{63.15}	&59.13	&61.27  \\ 
\cmidrule(lr){1-6}
\cmidrule(lr){7-11}
LR              &48.42	&53.60	&59.09	&52.51	&55.63   &52.08	&57.94	&62.88	&59.13	&60.65  \\
SVM	            &52.63	&51.54	&59.00	&53.08	&57.25	 &58.57	&57.78	&63.15	&58.12	&61.27  \\
NB	            &50.90	&53.32	&54.94	&52.78	&46.99	&55.62	&58.24	&55.09	&57.19	&50.13  \\ 
\cmidrule(lr){1-6}
\cmidrule(lr){7-11}

\textbf{Wikipedia}       &57.86	&\textbf{60.30}	&55.99	&59.19	&59.62  &64.16	&\textbf{66.83}	&60.94	&65.35	&66.41  \\ 
\cmidrule(lr){1-6}
\cmidrule(lr){7-11}
LR              &52.97	&55.90	&55.64	&56.40	&59.62  &55.68	&61.32	&60.16	&62.51	&66.41  \\
SVM	            &57.09	&57.48	&55.99	&57.15	&58.72	&64.16	&64.56	&60.94	&63.61	&64.81  \\
NB	            &57.86	&60.30	&51.64	&59.19	&57.29	&63.46	&66.83	&52.64	&65.35	&62.06  \\ 
\bottomrule
\end{tabular}
}
\label{tab:class-results-sc-all}
\end{table*}

SpinnerChief's settings for a stronger obfuscation of text (SpinnerChief-IF) increased the success rate with which the automated classification approaches identified the paraphrases. 
On average, SpinnerChief-DF replaced 12.58\% and Spinner-Chief-IF 19.37\% of the words in the text (see \Cref{sec:opt}). The 6.79\% increase in the number of replaced words for SpinnerChief-IF increased the average F1-Micro score of the classification approaches by 5.56\%. This correlation suggests the classification approaches can recognize most of the characteristic word replacements of the studied paraphrasing tools. 

The word choice and grammatical errors typical for texts of \textit{English as a second language} students decreased the classification performance less than we had expected. The SpinBot Theses test set's highest scores were $\approx$6\% lower than the highest scores for SpinBot overall. For SpinnerChief, the difference between the highest scores for Theses and overall was $\approx$2\%.

Furthermore, text-matching software and our classification approaches have complementary strengths regarding the detection of machine-paraphrased plagiarism. Text-matching software, such as Turnitin, is currently the de-facto standard technical support tool for identifying plagiarism. However, since these tools search for identical text matches, their detection effectiveness decreases when the number of replaced words increases \cite{WeberWulff19}. Including additional scans with the proposed classification approaches, as part of the detection process of text-matching software, could alleviate current systems' weaknesses.

Nevertheless, the F1-Micro scores for SpinnerChief-IF test cases were, on average, 13\% lower for SpinBot cases, although these cases exhibit a similar ratio of replaced words (see \Cref{tab:class-results-sc-all}). As for the SpinBot test cases, all approaches performed best for Wikipedia test cases and worst for Theses. However, the performance differences were smaller for the SpinnerChief test sets than for SpinBot test sets. For all SpinnerChief test sets, the lowest F1-Micro scores are at most 6.5\% below the highest scores for the test set, and the runner-ups are generally within an interval of 2\% of the best scores for the test sets.



Our transfer learning approach likely caused the drop in the classification performance and the overall leveling of the F1-Micro scores for SpinnerChief test cases. As explained in \Cref{sec:chap3_method_mpp}, we seek to provide a system with high generalization for different document collections and paraphrasing tools. Therefore, we used the machine-paraphrased text samples of SpinBot and applied the pre-trained word embedding models from \Cref{tab:mpp_word_embed_det} to extract the vector representations. We then used these vectors as features for the machine learning classifiers for both Spinbot and SpinnerChief test sets.


Considering the results combining machine learning techniques and word embedding features, for SpinBot (\Cref{tab:class-results-sb}) and SpinnerChief (\Cref{tab:class-results-sc-all}), we selected the top ranked ones to compose our \textit{Baseline} in the Transformer experiments on \Cref{tab:class-results-nlm}.

\subsubsection{Results for Transformer-based Architectures} \label{sec:eval_nlm}
The SpinBot test cases show all transformer models outperformed machine learning counterparts by significant margins, 16.1\% for Theses, and 13.27\% on average for all three data sources (\Cref{tab:class-results-nlm}). Our findings show all neural language models captured SpinBot's intrinsic paraphrasing method almost entirely. We stop the training for each model after one epoch to avoid overfitting to the paraphrasing method.

\begin{table*}
\caption[Classification results of neural Language Models against the best performing Machine Learning models on SpinBot and SpinnerChief.]{Classification results (F1-Micro) for SpinBot and SpinnerChief (DF and IF) using. The first block shows the best results from machine learning methods. The second block is composed of transformer-based optimization techniques, the third of new architectural or training approaches, and the fourth of our best-performing lexical \ac{KD} method. The highest scores are in \textbf{Bold} face.}
\centering
\resizebox{\textwidth}{!}{
\begin{threeparttable}
\begin{tabular}{l c ccc c ccc c ccc} 
\toprule
\multirow{2}{*}{Techniques} & {} & \multicolumn{3}{c}{\textbf{SpinBot}} & {} & \multicolumn{3}{c}{\textbf{SpinnerChief-DF}} & {} & \multicolumn{3}{c}{\textbf{SpinnerChief-IF}}  \\

\cmidrule(lr){3-5}
\cmidrule(lr){7-9}
\cmidrule(lr){11-13}
& {} & arXiv & Theses & Wiki  & {} & arXiv & Theses & Wiki & {} & arXiv & Theses & Wiki           \\
\toprule
Baseline                   & {} & 86.46\tnote{a} & 83.51\tnote{a}  & 89.55\tnote{a}     & {} & 59.78\tnote{b} & 59.09\tnote{c}  & 60.30\tnote{b}              & {} & 65.89\tnote{b} & 63.15\tnote{d}  & 66.83\tnote{b}               \\
\hdashline
\ac{BERT}                 & {} & 99.44 & 94.72  & 99.85      & {} & 50.74 & 50.42  & 43.00                & {} & 64.59 & 63.59  & 57.45               \\
\ac{ALBERT}                & {} & 98.91 & 96.77  & 99.54     & {} & 66.88 & 47.92  & 50.43             & {} & 75.57 & 56.75  & 59.61               \\
DistilBERT             & {} & 99.32 & 96.61  & 99.42      & {} & 38.37 & 45.07  & 37.05              & {} & 47.25 & 51.44  & 46.81               \\
\ac{RoBERTa}                     & {} & 99.05 & 97.34  & 99.85     & {} & 57.10  & 47.40 & 48.03             & {} & 66.00    & 58.24  & 58.94               \\
\hdashline
ELECTRA                   & {} & 99.20  & 96.85  & 99.41      & {} & 43.83 & 44.95  & 56.30               & {} & 60.77 & 63.11  & \textbf{75.92}               \\
BART                   & {} & 99.58 & 99.66  & 99.86      & {} & 69.38 & 53.39  & 48.62              & {} & 76.07 & 63.57  & 58.34               \\
XLNet                       & {} & \textbf{99.65} & 98.33  & 99.48      & {} & 69.90  & 53.06  & 50.51             & {} & 80.56 & 71.75  & 61.83               \\
Longformer                  & {} & 99.38 & \textbf{99.81}  & 99.87      & {} & \textbf{76.44} & \textbf{70.15}  & \textbf{63.03 }             & {} & 78.34 & 74.82  & 67.11               \\
\hdashline
\ac{LMGC-M} (XLNet) & {} & 99.48 & 98.86 & \textbf{99.90} & {} & 72.13 & 61.32 & 55.75 & {} & \textbf{82.32} & \textbf{75.23} & 61.23 \\
\bottomrule
\end{tabular}
\begin{tablenotes}[para]
    \item[a] \ac{GloVe}+\ac{SVM}
    \item[b] \ac{W2V}+\ac{NB}
    \item[c] \ac{D2V}+\ac{LR}
    \item[d] \ac{D2V}+\ac{SVM}
\end{tablenotes}
\end{threeparttable}
}
\label{tab:class-results-nlm}
\end{table*}

Longformer achieved the best performance on the SpinnerChief-DF test benchmark, increasing the F1-Micro score over machine learning baselines by up to 16.66\% for arXiv pre-prints, and 10.15\% on average for all three data sources. Longformer's architecture is designed to capture short and long dependencies with two main attention mechanisms: a sliding window and global attention. We assume the model generalizes well even when we decrease the word frequency with which words change to capture long dependencies better. However, comparing the results of SpinnerChief-DF in \Cref{tab:class-results-nlm}, we see a substantial drop in the F1 score for all approaches in relation to SpinBot and SpinnerChief-IF. Our hypothesis is the ratio of changing words of the paraphrased tools strongly affects our models.

When we increase SpinnerChief paraphrasing frequency to the same one as in SpinBot, most architectures significantly improve baseline performance. This observation is consistent with the discussion and results from machine learning experiments (see \Cref{tab:class-results-sb,tab:class-results-sc-all}). XLNet, Longformer, and ELECTRA achieve the highest results with an improvement of 14.67\%, 11.67\%, and 9.09\% in F1-score for arXiv, Theses, and Wikipedia, respectively. As ELECTRA was pre-trained using a Wikipedia Dump and the Books Corpus \cite{ZhuKZS15}, we assume it captured semantic aspects of Wikipedia articles as well. Additionally to Wikipedia, Longformer and XLNet also trained on other larger datasets, e.g., Gigaword 5 \cite{NapolesGV12}, CC Stories \cite{TrinhL19}, and Realnews \cite{ZellersHRB19}, which are seen more frequently during training. In return, Longformer and XLNet seem to capture unseen semantic structures better (e.g., arXiv and Theses) due to their high diversity in the training data.

When using XLNet as the backbone network for our best performing method (\ac{LMGC-M}) in \ac{WSD} and \ac{NLU}, we observe slight performance gains on average over XLNet. In the SpinnerChief-IF test case, we surpass the highest results for arXiv and Theses texts by 1.13\%. \ac{LMGC-M} generalizes well on the unknown spinning pattern, which is in line with our previous experiments in \ac{NLU} (\Cref{sec:chap4_eval_wsd}).
Considering the results of Longformer, we think it is a strong candidate for the backbone network in future versions of \ac{LMGC-M}.

The high compression rate of DistilBERT with a two-fold reduction in the number of parameters showed a significant decrease in accuracy over \ac{BERT} in all SpinnerChief test cases. Although we expected a slight decline in F1, the results exceeded our predictions. DistilBERT only performs 2.5\% worse than \ac{BERT} on the \ac{GLUE} \cite{WangSMH19} dataset, but it drops by 10.63\% F1-Micro on average on the SpinnerChief test cases, producing results comparable to random guessing. This result intrigued us how our converged Longformer$_{mini}$ model would perform in the task of \ac{PD} which we aim to address in future work.
\ac{ALBERT}'s parameter reduction techniques (e.g., factorized embedding parametrization, parameter sharing) seem to be more robust and outperform \ac{BERT} with a 4.56\% increase on average on SpinnerChief test cases.

\ac{RoBERTa} performs slightly better than \ac{BERT}, with an improvement of 0.99\% on average. As \ac{RoBERTa} uses more parameters than most other \ac{BERT}-related models and has exceptionally high computational requirements for pre-training, we rate this advance as negligible.
In most SpinnerChief test cases, the \ac{BERT}-based approaches performed comparably to the machine learning baselines. Models that are strongly related to \ac{BERT} are not performing as well on the paraphrase detection task as models that significantly change the attention scheme or training objective.

Given the evidence at hand, spun words' frequency is a strong indicator of our models' performance. However, since the spinning method of SpinBot and SpinnerChief (DF and IF) is unknown and could be different for each case, we can interpret the findings in two ways. First, the models may capture spinning frequency intrinsically and increase their attention to more words, which would mean the methods can better detect highly spun paragraphs. Second, the spinning method in SpinnerChief-IF can be akin to the one used in SpinBot. 

At a higher level, the techniques proposing different training architectures performed better than those based on hyperparameter optimization from \ac{BERT}. We believe the windowed local-global self-attention scheme used in Longformer allowed the model to generalize better between different paraphrasing tools. In 8 out of 9 scenarios, Longformer was either the best or second-best model overall. Also, for almost all cases, the neural language approaches surpassed the machine learning ones, providing an attractive solution for the paraphrased detection problem. Concluding, we can see the results on SpinnerChief-DF as a lower bound for unseen spinning methods, even if the frequency of word replacements is drastically changed. 

\subsection{Human Baseline} \label{sec:eval_human}
To complement our earlier study \cite{FoltynekRSM20}, we conducted a user survey with excerpts from ten randomly selected Wikipedia articles. We paraphrased three excerpts using SpinerChief-DF and SpinnerChief-IF and kept four excerpts unaltered.

Using QuizMaker\footnote{\url{https://www.quiz-maker.com/}}, we prepared a web-based quiz showing the ten excerpts one at a time and asked the participants to vote whether the text had been machine-paraphrased. We shared the quiz via e-mail and a Facebook group with researchers from the academic integrity community. We obtained 32 responses. The accuracy of the participants ranged between 20\% and 100\%, with an average of 65.59\%. Thus, the F1-Micro score of the average human examiner matched the average of the best scores for SpinnerChie-IF test sets (65.29\%). Some participants pointed out some excerpts' irregularities, e.g., lowercase letters in acronyms, helped them identify the excerpts as paraphrased. For SpinBot, which we evaluated in our preliminary study, 73 participants answered the survey with an accuracy between 40\% and 100\% (avg. 78.40\%) \cite{FoltynekRSM20}.  

Our experiments show that experienced educators who read carefully and expect to encounter machine-paraphrased text could achieve accuracy between 80\% and 100\%. However, even in this setting, the average accuracy was below 80\% for SpinBot and below 70\% for SpinnerChief. We expect the efficacy will be lower in realistic scenarios, when readers focus less on spotting machine-paraphrased text. 

\subsection{Text-matching Software Baseline} \label{sec:eval_turnitin}
To quantify the benefit of our automated classification over text-matching software, we tested how accurately current text-matching tools identify paraphrased text. We used two systems - \mbox{Turnitin}\footnote{\url{https://www.turnitin.com/}}, which has the largest market share, and \mbox{PlagScan}\footnote{\url{https://www.plagscan.com/}} - one of the best-performing systems according to a test conducted by the European Network for Academic Integrity \cite{FoltynekDAR20}. We created 160 documents to simulate patch-writing - 2x40 from Wikipedia, 40 from arXiv, and 40 from Theses. Each document contains 20 randomly chosen paragraphs. Each set of 40 documents contains 10x4 files.
The first dimension captures the paragraph length in sentences (varying from 1 to 10), and the second dimension represents the spinning tool (original, SpinBot, SpinnerChief-DF, and SpinnerChief-IF).

To ensure the text matching-software test is objective and comparable across the data sets, we used solely the overall percentages of text-match reported by a system. In most cases, systems correctly identified the right source. However, in many cases, there were also false positives caused by random matches, which means the systems' actual success is slightly lower than reported in the tables.

The results (see \Cref{tab:results-plag-turn}) show PlagScan struggles in the scenario of patch-writing. Even though the \ac{PD} tool indexes Wikipedia and identifies plagiarism of the whole documents \cite{FoltynekDAR20}, the average text match reported for patch-written documents was 63\%. Spinning with SpinBot and SpinnerChief-IF reliably prevented PlagScan from detection. For SpinBot, the average reported percentage of text-match was only 1\%, and for SpinnerChief-IF 3\%. In the case of SpinnerChief-DF PlagScan managed to identify 19\% of plagiarism, which is caused by less portion of altered words. Nonetheless, considering the ground truth for all documents of 100\%, we can conclude that patch-writing combined with spinning successfully avoids detection by PlagScan.

\begin{table}[htb]
\caption[Classification results of two Plagiarism Detection systems: Turnitin and PlagScan]{Classification results (text-match in \%) of two Plagiarism Detection systems: Turnitin and PlagScan.}
\centering
\resizebox{.85\textwidth}{!}{
\begin{tabular}{lccccccccc}\toprule
\multirow{2}{*}{} &\multicolumn{3}{c}{\textbf{Turnitin}} & \multicolumn{3}{c}{\textbf{PlagScan}}  \\
\cmidrule(lr){2-4}
\cmidrule(lr){5-8}
                & arXiv    & Theses      &  Wikipedia        
                & arXiv    & Theses      &  Wikipedia    \\
\midrule

Original                 &84.0	&5.4	&98.7   &44.6	&22.3	&65.0 \\
SpinBot                  &7.0	&1.1	&30.2	   &0.0	&0.1	&0.5	      \\
SpinnerChief-DF          &58.5	&4.0	&74.5	  &9.2	&12.0	&19.1		 \\
SpinnerChief-IF          &38.8	&1.2	&50.1   &1.8	&0.5	&3.1	 	 \\

\bottomrule
\end{tabular}
}
\label{tab:results-plag-turn}
\end{table}

As shown in \Cref{tab:results-plag-turn}, Turnitin performed better on patch-written documents than PlagScan. For Wikipedia, Turnitin reported 100\% text-match in almost all cases. Also, the text-match reported for spun documents was much higher than for PlagScan --- 31\% for SpinBot, 74\% for SpinnerChief-DF, and 50\% for SpinnerChief-IF. Nonetheless, text-spinning prevents Turnitin from identifying a significant portion of plagiarism. Whereas Turnitin copes better with patch-writing, it does not index as many Theses as PlagScan.

For both systems, we observe longer fragments lead to higher identification rates. This result corresponds without classification, which also yielded higher accuracy for longer passages.

This experiment extended our previous tests \cite{FoltynekRSM20}, in which the documents contained unaltered fragments interspersed with three paraphrased ones. Turnitin identified all unaltered fragments and one of the machine-paraphrased fragments. Of the other two machine-paraphrased fragments, Turnitin incorrectly credited one to another source and missed the other. These results were in line with the findings of \newcite{RogersonM17}, who used two paraphrasing tools to obfuscate a paragraph from a prior publication. 
When Turnitin received original paragraphs as inputs, the system found a 100\% match with the source. However, for the two machine-paraphrased versions of the paragraph, Turnitin computed a similarity score of zero.

From these experiments, we conclude if plagiarists copy a few paragraphs and employ a paraphrasing tool to obfuscate their misconduct by spinning, the similarity is often below the text-matching tool's threshold, thus causing the plagiarism to remain undetected. Classification of machine-paraphrased text may be a useful complement of standard text-matching, which can alert users when there is a suspicion of deliberate obfuscation of plagiarism.

\subsection{Limitations}
\ac{ML} techniques can supplement text-matching software to detect plagiarism from unknown sources and in cases where spun paragraphs do not achieve the \ac{PD} system's threshold. However, the presented classification techniques provide a classification score with no intuitive interpretation (e.g., similarity to a document, similarity in writing style to an author). Therefore, positive classifications require additional manual validation of the affected paragraphs to avoid misleading plagiarism allegations resulting from false positives.

\chapter{Final Considerations} \label{Chapter5}

This thesis addressed the problem of increasing \ac{LM} size, which we see as a risk to the research field. Large model sizes hinder research groups, without access to necessary hardware infrastructures, from experimenting with custom pre-trained models. Furthermore, an increased model size leads to increases in energy consumption and makes on-device deployment difficult.
Our presented methods counteract these problems with an efficient architecture, gaining multiple teachers' knowledge and leveraging lexical knowledge sources.
This chapter presents the final considerations to this thesis. Starting with a conclusion of the experiments (\Cref{sec:chap5_conclusion}), we state a broader impact of our research contributions (\Cref{sec:chap5_broader_impact}). Finally, we outline our future work in the domains of \ac*{KD}, \ac{WSD}, and \ac{PD} (\Cref{sec:chap5_future_work}).

\section{Conclusion} \label{sec:chap5_conclusion}

We introduced a method to perform \ac{KD} using the knowledge of four large-scale \ac{LM}s. Our method weights predictions of each teacher by confidence and uses them as targets for our efficient architecture (Longformer$_{mini}$). During training, the student model achieved faster convergence in language modeling than its larger counterparts, although using fewer parameters.
However, the experiments conducted in this work did not provide conclusive answers yet. Thus we focussed on lexical \ac{KD} solutions, which appeared more promising. 

We proposed two methods to perform \ac{WSD} (\ac{LMGC} and \ac{LMGC-M}) with eight neural \ac{LM}s. Our methods distill knowledge from \ac{LKB} which proved to be beneficial for \ac{NLU} tasks (e.g., text-similarity). \ac{LMGC} and \ac{LMGC-M} perform \ac{WSD} by combining neural \ac{LM}s with lexical resources from WordNet. We exceeded state-of-the-art \ac{WSD} methods on five representative datasets (+0.5\%) and improved the performance over \ac{BERT} in general language understanding tasks (+1.1\%). We release the code\footnote{\url{https://github.com/jpwahle-word-sense-disambiguation/wsd}}, and pre-trained models\footnote{\label{footnote:huggingface_chap5}\url{https://huggingface.co/models?search=jpwahle}} to reconstruct our experiments.

The \ac{MPP} detection experiments reported in this thesis extended a preliminary study \cite{FoltynekRSM20} by analyzing two new collections (arXiv, Theses), a new paraphrasing tool (SpinnerChief), and eight neural language models based on the transformer architecture against our best-performing method (\ac{LMGC-M}). We selected training and test sets reflecting documents particularly relevant for the plagiarism detection use case. The arXiv collection represents scientific papers written by expert researchers. Graduation Theses of non-native English speakers provide writing samples of authors whose style varies considerably. Wikipedia articles represent collaboratively authored documents for many topics and one of the sources from which students plagiarize most frequently.  
The classification approaches we devised are robust to identifying machine-paraphrased text, which educators face regularly. To support practitioners and facilitate an extension of the research on this important task, the data\footnote{\url{https://doi.org/10.5281/zenodo.3608000}}, code\footnote{\url{https://github.com/jpwahle/iconf22-paraphrase}}, and pre-trained models\textsuperscript{\ref{footnote:huggingface_chap5}} of our study, as well as the web-based demonstration system are openly available\footnote{\url{http://purl.org/spindetector}}.

\section{Broader Impact} \label{sec:chap5_broader_impact}
Our work focussed on distilling the knowledge of transformer language models and lexical resources to increase semantic representations. We see a potential positive impact with our methods to increase the accessibility for research purposes. Our model contributes to creating high-performing models at no additional hardware overhead, which benefits small research groups. 
Furthermore, as we have shown, Longformer$_{mini}$ uses an efficient attention scheme and a lower number of parameters than other \ac{LM}s, resulting in less computational expenses when using the trained model. These computational expenses can translate into greenhouse gas emissions \cite{StrubbellGM19} which would reduce when organizations and institutions use Longformer$_{mini}$ for deployment. 

The lexical \ac{KD} methods presented in this thesis successfully gained semantic understanding about polysemous words, which transferred to tasks beyond \ac{WSD}. We think polysemy resolution will become crucial to broader domains such as the processing of mathematical formulas. Mathematical documents communicate their information in an ambiguous, context-dependent, and non-formal language \cite{GreinerPetterRSA19}. For example, one equation might have a possibly infinite amount of equations with identical meaning.
Our methods could disambiguate mathematical equations for information retrieval applications using external knowledge (e.g., using a comprehensive typology for mathematical notations, similar to WordNet for polysemous words).

We think the \ac{LM} research field focusses more on improving task performance than questioning model size and computational expenses. Therefore researchers using smaller models may, at first, not be competitive against large models running on large hardware infrastructures. We believe with more publications about efficient \ac{LM}s and the impact of large-scale models; researchers will weight these effects as important as task performance.
Aside from this concern, we see no negative ethical or societal impacts of our work beyond what also applies to other core components of \ac{LM}s.

\section{Future Work} \label{sec:chap5_future_work}

We aim to extend our multi-teacher \ac{KD} experiments with converging Longformer$_{mini}$'s training fully. We think the converged student model could achieve comparable performance to its teachers and other \ac{KD} techniques on general \ac{NLU} tasks. The \ac{GLUE} benchmarks is a logical choice as it allows us to compare the model to our lexical \ac{KD} methods.

With an increasing amount of transformer-based language models proposed in short amounts of time, the choice for \ac{KD} teachers grows too. This thesis chose teachers by training objective, vocabulary, and the datasets used for training. We expected prediction accuracy to be high when the datasets were similar as teacher models already trained on similar data. However, another perspective would be to use teachers trained on very different datasets to capture knowledge from many domains. This experiment may ask for a higher penalty for unconfident predictions than we used in our method as some teachers might not produce meaningful predictions.

Our future work in the domain of \ac{WSD} will include testing generalization on the WiC \cite{PilehvarC19} and SuperGLUE \cite{WangPNS19} datasets. Furthermore, we want to test discriminative fine-tuning \cite{HowardR18} against our parallel approach, i.e., whether training \ac{MLM} on the \ac{LKB} first, and subsequently optimizing for \ac{WSD} yields the same effect as \ac{LMGC-M}.

The lexical \ac{KD} methods use multiple components to increase the model's semantic understanding. In the future, we seek to investigate which components of our methods are the most impacting by performing a more detailed ablation study. We conducted preliminary experiments with other knowledge bases (e.g., Wikidata) but leave for future work to incorporate knowledge from these sources. 

We intend to extend the experimental results from \ac{WSD} experiments to \ac{NLU} benchmarks using the trained \ac{LMGC} and \ac{LMGC-M} models with XLNet. Although \ac{LMGC-M} showed superior performance on average in the \ac{GLUE}, we aim to estimate its classification impact on polysemy. 

The \ac{MPP} detection experiments indicated that obtaining additional training data is a promising approach for improving \ac{ML} backed approaches for identifying machine-paraphrased text. Additional training data should cover more paraphrasing tools, topics, and languages. We see a community-driven open data effort (or crowdsourcing) as a promising option for generating a comprehensive training set.


Another interesting direction would be to use \ac{AE} or \ac{AR} to spin words or generate new text. This setup will be more realistic in the future as \ac{LM}s are publicly available and generate texts that are difficult to distinguish from human writing.

Considering Longformer's superior performance in most tasks and the improvement of \ac{LMGC-M} over its backbone network (XLNet), we see two interesting future experiments with our proposed methods. First, using Longformer as the backbone network for \ac{LMGC-M}, and second, applying the converged Longformer$_{mini}$ model to the task of \ac{PD}. 

Some of these ideas have been realized in later works built on this thesis such as \ac{AE} \cite{Wahle21b} and \ac{AR} paraphrasing \cite{Wahle2022d}. Other works have used concepts from this thesis for applications such as abstractive text summarization \cite{Kirstein2022} or misinformation detection about COVID-19 \cite{10.1007/978-3-030-96957-8_33}. 

\bibliographystyle{apalike}
\bibliography{references.bib} 


\appendix 

\chapter{Dataset Details} \label{AppendixA}

As our experiments in \Cref{sec:chap4_eval_wsd} showed, the additional learning of \ac{WSD} in parallel to \ac{MLM} leverages our model's performance to other natural language understanding tasks. One direct result of incorporating \ac{LMGC} training methods is the increased results on the \ac{GLUE} tasks. To investigate the correlation between the number of polysemous words and \ac{LMGC}, we report how many words from each \ac{GLUE} tasks can be found as polysemous in WordNet, as Table \ref{table:glue_polysemy} shows.

\begin{table*}[htb]
\caption[Polysemy in the GLUE evaluation datasets]{The number of polysemous words in relation to the number of total words in the \ac{GLUE} evaluation datasets except for WNLI. Under each dataset is the number of training examples.}
\centering
 \resizebox{\textwidth}{!}{
    \begin{tabular}{llcccccccc} \toprule
     & & MNLI-(m/mm) & QQP & QNLI & SST-2 & CoLA & STS-B & MRPC & RTE \\
     & & 392k & 363k & 108k & 67k & 8.5k & 5.7k & 3.5k & 2.5k \\ \midrule
    \multirow{3}{*}{Train} & Words & 11\,695k & 8\,050k & 3\,818k & 633k & 65k & 114k & 160k & 130k \\
    {} & Polysemous & 5\,205k & 3\,452k & 1\,439 & 366k & 23k & 51k & 65k & 50k \\
    {} & Relation (\%) & 44.5& 42.9 & 37.7 & 57.8 & 35.5 & 44.9 & 40.9 & 39.0 \\ \midrule
    \multirow{3}{*}{Test} & Words & 286k & 893k & 205k & 17k & 8k & 34k & 17k & 14k \\
    {} & Polysemous & 126k	& 382k & 78k & 9k & 2k & 15k & 7k & 5k \\
    {} & Relation (\%) & 44.2 & 42.8 & 8.2 & 54.1 & 34.7 & 46.4 & 41.9 & 38.7 \\
    \bottomrule
    \end{tabular}}
\label{table:glue_polysemy}
\end{table*}

Our first observation is the number of polysemous words in the text from different \ac{NLU} tasks is high in general. We find specific tasks like QQP with 42.8\% polysemous words in the test set with a high accuracy boost in \ac{LMGC-M}, but other tasks like SST-2 with 54.1\% in the test set with almost no improvement. This result indicates \ac{LMGC-M} can improve the performance with an increased number of polysemous words but does not necessarily mean there is a causation. The baseline model \ac{BERT}$_{base}$ already learns \ac{WSD} to some extent to resolve polysemy in text, which the results in \ac{GLUE} tasks show. Also, the difficulty of some tasks is not bound to polysemous words. However, as our experimental results showed, on average, the performance of \ac{LMGC-M} is higher than the original \ac{BERT}$_{base}$ model, which indicates \ac{WSD} influence general language understanding.

\end{document}